\documentclass{article}

\usepackage[preprint,final,nonatbib]{neurips_2025}

\usepackage[utf8]{inputenc}
\usepackage[T1]{fontenc}
\usepackage{hyperref}
\usepackage{url}
\usepackage{booktabs}
\usepackage{amsfonts}
\usepackage{amssymb}
\usepackage{mathtools}
\usepackage{amsmath}
\usepackage{nicefrac}
\usepackage{microtype}
\usepackage[table, dvipsnames]{xcolor}  %
\definecolor{darkgrey}{gray}{0.3}
\usepackage{graphicx}
\usepackage{multirow}
\usepackage{mdframed}
\hypersetup{
  colorlinks=true,
  linkcolor=magenta,       %
  citecolor=RoyalBlue,      %
  urlcolor=gray      %
}
\usepackage[backend=biber,style=authoryear-comp,sorting=nyt,sortcites=true,maxnames=2,hyperref=true, backref=true]{biblatex}

\newmdtheoremenv[
  linecolor=darkgrey,
  linewidth=2pt,
  innertopmargin=6pt,
  innerbottommargin=6pt,
  skipabove=10pt,
  skipbelow=10pt
]{definition}{Definition}[section]

\title{Topotein: Topological Deep Learning for Protein Representation Learning}

\author{%
  Zhiyu Wang \\
  University of Cambridge \\
  Cambridge, UK \\
  \texttt{zw471@cam.ac.uk} \And
  Arian Jamasb \\
  Prescient Design \\
  Basel, Switzerland \\
  \texttt{arian@jamasb.io}\And
  Mustafa Hajij \\
  University of San Francisco \\
  San Francisco, California, USA \\
  \texttt{mhajij@usfca.edu} \And
  Alex Morehead \\
  Lawrence Berkeley National Laboratory \\
  Berkeley, California, USA \\
  \texttt{acmwhb@lbl.gov} \And
  Luke Braithwaite \\
  University of Cambridge \\
  Cambridge, UK \\
  \texttt{lb2027@cam.ac.uk} \And
  Pietro Liò \\
  University of Cambridge \\
  Cambridge, UK \\
  \texttt{pl219@cam.ac.uk} \\
}

\begin{document}

\maketitle
\begin{abstract}
  Protein representation learning (PRL) is crucial for understanding structure-function relationships, yet current sequence- and graph-based methods fail to capture the hierarchical organization inherent in protein structures. We introduce Topotein, a comprehensive framework that applies topological deep learning to PRL through the novel Protein Combinatorial Complex (PCC) and Topology-Complete Perceptron Network (TCPNet). Our PCC represents proteins at multiple hierarchical levels---from residues to secondary structures to complete proteins---while preserving geometric information at each level. TCPNet employs SE(3)-equivariant message passing across these hierarchical structures, enabling more effective capture of multi-scale structural patterns. Through extensive experiments on four PRL tasks, TCPNet consistently outperforms state-of-the-art geometric graph neural networks. Our approach demonstrates particular strength in tasks such as fold classification which require understanding of secondary structure arrangements, validating the importance of hierarchical topological features for protein analysis. Our code is made available at \href{https://github.com/ZW471/TopoteinWorkshop}{github.com/ZW471/TopoteinWorkshop} for reproduction.
\end{abstract}

\section{Introduction}\label{sec:introduction}
Advances in protein structure prediction have revolutionized structural data availability~\parencite{jumperHighlyAccurateProtein2021, linEvolutionaryscalePredictionAtomiclevel2023}, with over 200 million protein structures now available through AlphaFold Database~\parencite{varadiAlphaFoldProteinStructure2022} and even larger collections like ESM Atlas~\parencite{linEvolutionaryscalePredictionAtomiclevel2023} containing over 600 million structures. However, these structures remain largely unannotated, creating a critical gap between available data and functional understanding~\parencite{jamasbEvaluatingRepresentationLearning2024}. Protein Representation Learning (PRL) addresses this challenge by learning vector representations that bypass manual feature engineering, automatically extracting meaningful patterns from protein sequences or structures to enable effective downstream prediction of functional properties~\parencite{gligorijevicStructurebasedProteinFunction2021a}.

Current PRL approaches fall into two main categories: transformer-based protein language models that process amino acid sequences~\parencite{linEvolutionaryscalePredictionAtomiclevel2023, heinzingerBilingualLanguageModel2024, elnaggarProtTransCrackingLanguage2020, elnaggarAnkhOptimizedProtein2023}, and structure-based geometric graph neural networks (GGNNs) that represent proteins as 3D graphs of residues~\parencite{moreheadGeometrycompletePerceptronNetworks2024, jingLearningProteinStructure2020, zhangProteinRepresentationLearning2022}. However, both approaches are typically limited to learning local representations at the residue level. They fundamentally miss the hierarchical organization of proteins that would enable learning representations simultaneously across multiple levels, including the more global secondary structure level.

A holistic understanding of proteins requires learning representations at all hierarchical levels simultaneously. In particular, secondary structures and their spatial arrangements are central to how biologists understand and classify proteins. Popular protein structure classification systems like CATH and SCOP employ hierarchical schemes where three out of four levels in CATH and half the levels in SCOP are based on secondary structure organization~\parencite{knudsenTheCATHDatabase2010, murzinSCOPStructuralClassification1995}. This biological importance translates directly to computational applications: recent generative modeling work like ProtComposer~\parencite{starkProtComposerCompositionalProtein2025} has demonstrated that using 3D ellipsoids that represent secondary structures to guide the protein generation process significantly improves protein design quality, achieving better designability, novelty, and structural diversity compared to residue-only approaches. The success in generative modeling suggests that secondary structure representations could similarly enhance PRL when combined with residue-level representations, enabling more expressive pure-structure representations that better capture the hierarchical organization of protein architecture. This can be particularly important for applications such as more expressive pre-sequence filtering and inverse folding in de novo antibody design.

\begin{figure}[t]
    \centering
    \includegraphics[width=\textwidth]{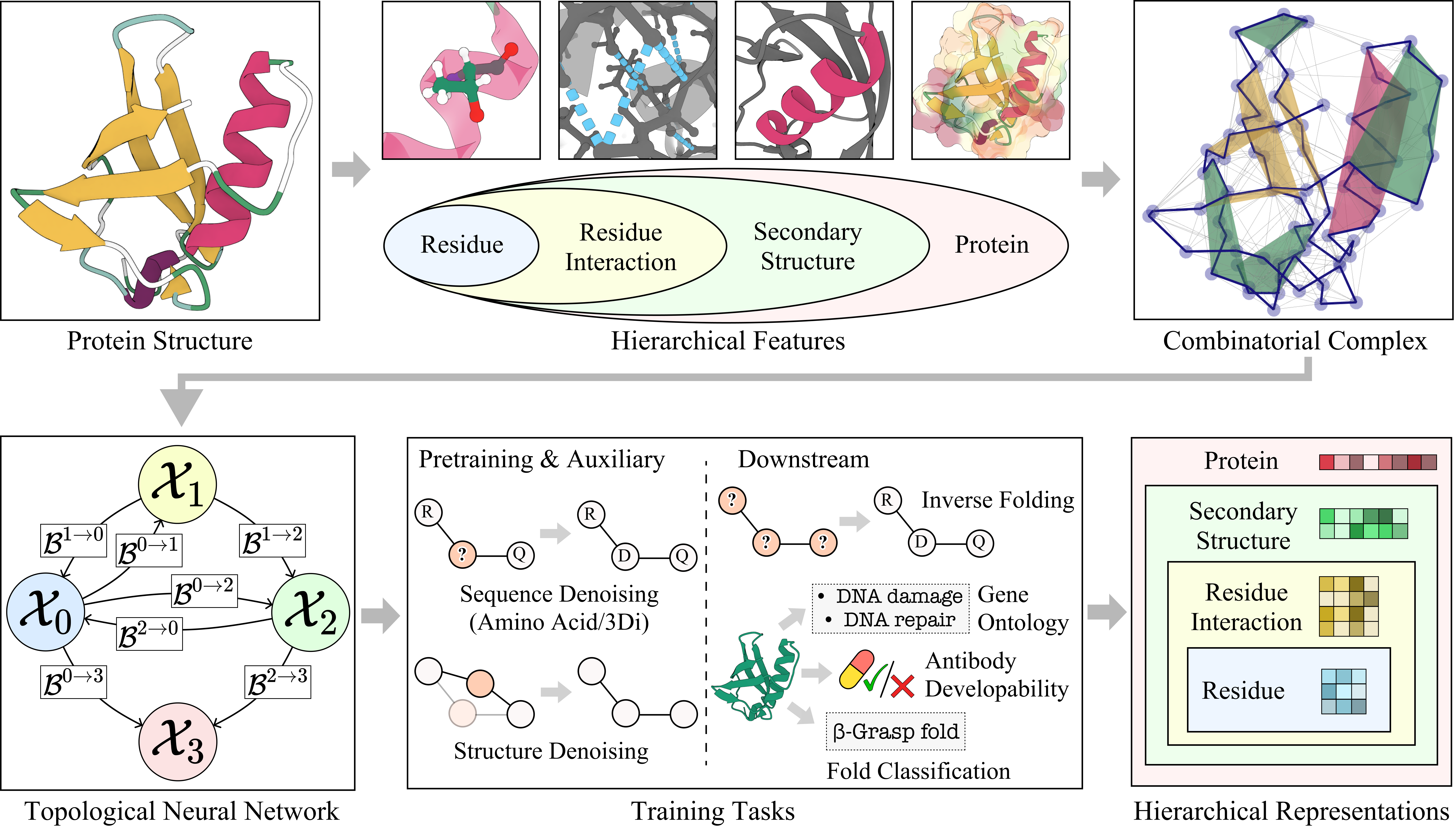}
    \caption{\textbf{Overview of the \textit{Topotein} framework.} Given PDB protein structures~\parencite{bermanProteinDataBank2000}, we construct \textit{Protein Combinatorial Complexes} that hierarchically organize residues (rank-0), interactions (rank-1), secondary structures (rank-2), and complete proteins (rank-3). These multi-rank representations can be processed by any topological neural network (e.g. our TCPNet) to generate protein embeddings for downstream prediction tasks.}
    \label{fig:visual_abstract}
\end{figure}

Topological Deep Learning (TDL)~\parencite{papillonArchitecturesTopologicalDeep2023} offers a promising approach to address these hierarchical modeling limitations by extending graph neural networks to generalized topological domains. Among these domains, the most popular is hypergraphs, which allow edges to connect sets of nodes rather than pairs, enabling higher-order relationships but still representing flat structures. On the other hand, simplicial and cellular complexes address hierarchical modeling by representing relationships beyond node and edge levels---from nodes to edges to faces and volumes. However, these impose strict boundary requirements where a rank-$n$ cell ($n$-cell for short) can only exist if all boundary $(n-1)$-cells also exist, introducing artificial constraints for biological systems.

A recently introduced TDL domain, combinatorial complexes~\parencite{hajijTopologicalDeepLearning2022}, provides a more flexible solution by generalizing both cellular complexes and hypergraphs, merging the hierarchical properties of cellular complexes with the flexible set-type relationships of hypergraphs while removing strict boundary constraints. One of the most representative works using combinatorial complexes is E(n)-equivariant Topological Neural Networks (ETNN)~\parencite{battiloroEquivariantTopologicalNeural2024}, which achieved state-of-the-art results on small molecule property prediction by representing molecules as combinatorial complexes with atoms as rank-0 cells, bonds as rank-1 cells, and rings/functional groups as rank-2 cells. This flexibility makes combinatorial complexes particularly suitable for protein representations, where secondary structures can be modeled as higher-order cells containing residues without artificial edges. However, ETNN has not been tested on large biomolecules like proteins and lacks the support for geometric features such as orientations that are needed for protein secondary structures.
This raises our central research question: \textit{What novel topological data structures and neural architectures are needed to explicitly capture protein-specific hierarchical inductive biases for more effective protein representation learning?}

We answer this question by presenting \textbf{Topotein} (Figure~\ref{fig:visual_abstract}), the first topological framework for learning protein structures. Following a standard PRL workflow~\parencite{jamasbEvaluatingRepresentationLearning2024}, we introduce two key innovations: Protein Combinatorial Complexes (PCCs) as a hierarchical set-type structure to replace traditional pair-wise graphs, and topological neural network (TNN) encoders such as TCPNet that leverage these multi-rank representations to generate protein embeddings for downstream tasks. Our contributions are threefold:

\begin{itemize}
    \item \textbf{Protein Combinatorial Complex}: A hierarchical data structure representing proteins at multiple levels with comprehensive featurization schemes for scalar and vector features, along with localized reference frames enabling SE(3)-equivariant topological networks.

    \item \textbf{Topology-Complete Perceptron Network (TCPNet)}: An SE(3)-equivariant TNN designed for hierarchical protein structures, featuring four-step hierarchical message passing that leverages the multi-rank representations and localized reference frames from PCCs.

    \item \textbf{Comprehensive Experimental Validation}: We demonstrate that TCPNet consistently outperforms state-of-the-art GNNs across four protein analysis tasks, with improved performance and robustness in structure-only scenarios. Additionally, we show that effective TDL requires careful architectural design by comparing against adapted versions of existing methods (GVP-GNN~\parencite{jingLearningProteinStructure2020} and ETNN~\parencite{battiloroEquivariantTopologicalNeural2024}).
\end{itemize}

\section{Why Topological Representations for Proteins?}\label{sec:why-topological}

\begin{figure}[hptb]
\centering
\includegraphics[width=\textwidth]{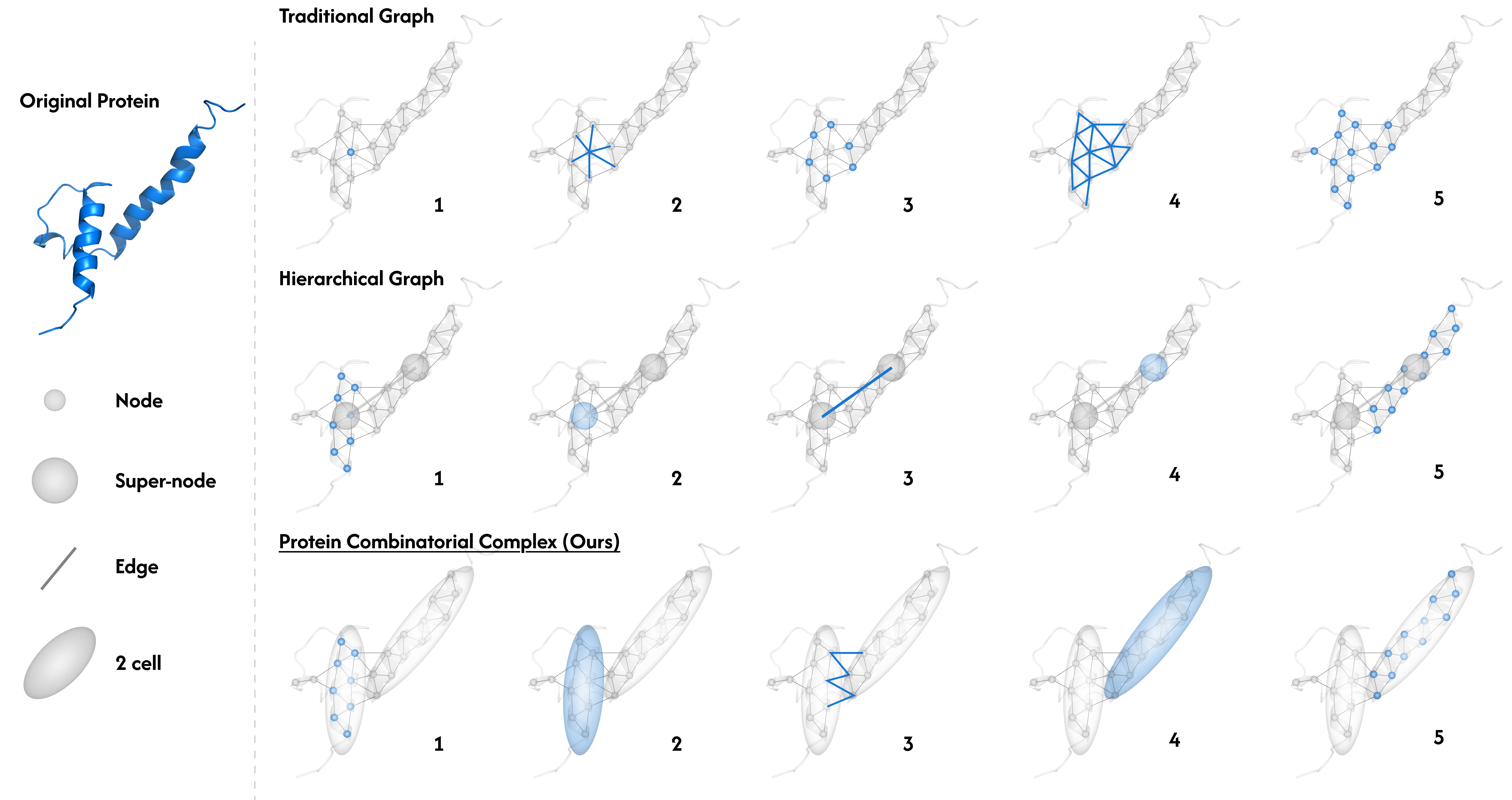}
\caption{\textbf{Message passing comparison across protein representation paradigms.} The figure illustrates three approaches to protein representation, each showing five consecutive steps of passing information from one secondary structure to another. Note that there should also be edges between nodes and supernodes in the middle row, but they are omitted for clarity. }
\label{fig:ccc_message_passing}
\end{figure}

Traditional GNNs represent proteins as graphs where nodes are residues and edges connect spatially proximate residues, typically determined by k-nearest neighbors or radius cutoffs.~\parencite{huAdvancesDeepLearning2024} While effective for many tasks, this representation creates fundamental bottlenecks for protein modeling. As shown in the top row of Figure~\ref{fig:ccc_message_passing}, traditional protein graphs suffer from severe information bottlenecks. Most messages are passed within individual secondary structure elements (SSEs) with very limited communication between them. Even worse, stacking multiple GNN layers causes information to echo repeatedly within the same SSE while still failing to effectively reach nearby SSEs, creating inefficient learning dynamics. One potential solution uses heterogeneous graphs with supernodes representing SSEs and superedges to connect them (middle row of Figure~\ref{fig:ccc_message_passing}). While this partially addresses connectivity issues by enabling direct communication between SSEs, it loses critical geometric information about SSE shapes and orientations---$\alpha$-helices lose their rod-like geometry and $\beta$-sheets lose their planar arrangement. Moreover, this method cannot model the precise contacts between SSEs, as the edges connecting them can only model simple spatial relationships such as the displacement between their center of mass.

The bottom row of Figure~\ref{fig:ccc_message_passing} demonstrates how TDL addresses both limitations through our combinatorial complex approach. Notice that in step 3, we can still use edges for message passing rather than superedges, utilizing multiple edges simultaneously for each pair of SSEs. Each edge contains its own information about the spatial relationship between the two SSEs. As a result, no geometric information is lost while enabling efficient hierarchical communication. To understand the theoretical foundations of TDL and the specific combinatorial complex framework that enables these advantages, we provide detailed background on TDL domains and neural architectures in the following section, with our approach detailed in Section~\ref{subsec:protein-combinatorial-complex}.

\section{Topological Deep Learning}\label{sec:background}
\subsection{Topological Domains}\label{subsec:domains}

\begin{figure}
    \centering
    \includegraphics[width=\textwidth]{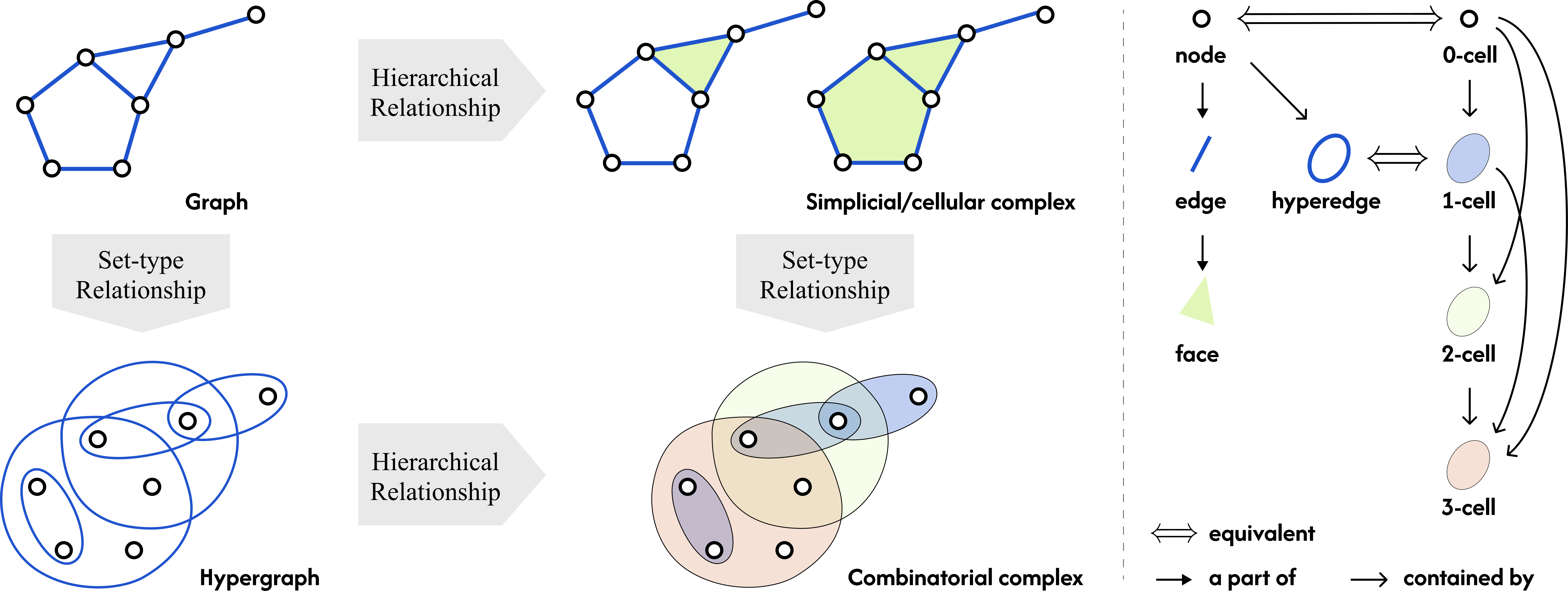}
    \caption{\textbf{Domains of Topological Deep Learning.} The three main topological domains: hypergraphs (left bottom) with hyperedges connecting multiple nodes; simplicial/cellular complexes (center top) with hierarchical structure and strict boundary requirements; combinatorial complexes (center bottom) unifying both approaches without boundary constraints. The right panel shows relationship types between cells of different ranks: \textit{part of} arrows indicate how lower rank cells combine to form higher rank cells (e.g., three edges forming a triangular face), \textit{contained by} arrows show how lower rank cells can be contained within higher rank cells without being structural components (e.g., a 3-cell may contain 1-cells but no 2-cells), and \textit{equivalence} arrows illustrate how traditional graph elements map to cells of different ranks in combinatorial complexes}
    \label{fig:tdl-domains}
\end{figure}

As shown in Figure~\ref{fig:tdl-domains}, TDL~\parencite{papillonArchitecturesTopologicalDeep2023, hajijTopologicalDeepLearning2022} extends geometric deep learning by generalizing graphs to topological domains that can capture complex relationships beyond simple pairwise connections. Understanding these domains is crucial for selecting the appropriate representation for protein modeling. For readers new to TDL, we provide an introductory primer in Appendix~\ref{sec:tdl-primer}. To appreciate this generalization, recall that a standard graph $\mathcal{G} = (\mathcal{V}, \mathcal{E})$ consists of a finite node set $\mathcal{V} = \{v_0, v_1, \ldots, v_n\}$ and edge set $\mathcal{E} = \{e_0, e_1, \ldots, e_m\}$, where each edge $e_{ij} = (v_i, v_j)$ represents a pairwise relationship. TDL generalizes this concept along two complementary directions: (1) extending to \emph{hierarchical relationships} through complexes that organize elements into part-whole structures, and (2) extending to \emph{higher-order relationships} through hypergraphs that allow interactions among arbitrary numbers of entities.

\paragraph{Hypergraphs.} Hypergraphs extend standard graphs by replacing pairwise edges $e_{ij} = (v_i, v_j)$ with hyperedges $e_{\{i,j,\ldots,k\}} = \{v_i, v_j, \ldots, v_k\}$ that can connect arbitrary numbers of nodes simultaneously, employing a two-step message-passing scheme where hyperedges first aggregate information from all contained nodes, then broadcast this aggregated information back to all member nodes~\parencite{heydariMessagePassingNeural2022}. However, hypergraphs have a fundamental limitation for protein modeling as they capture set-type relationships but lack hierarchical structure, unable to represent the nested organization where residues belong to secondary structures, which in turn belong to domains and complete proteins.

\paragraph{Simplicial/cellular complexes.} Simplicial and cellular complexes address this hierarchical limitation by organizing elements into part-whole relationships with strict boundary constraints, where nodes (0-cells) combine to form edges (1-cells), edges combine to form faces (2-cells), and so on. The key characteristic is the boundary requirement: a higher-rank cell can only exist if all its lower-rank boundary cells also exist. While this hierarchical structure is appealing for protein modeling, the strict boundary requirements create significant challenges as proteins exhibit complex structural arrangements where secondary structures may have irregular boundaries or variable lengths that cannot easily conform to rigid geometric constraints.

\paragraph{Combinatorial Complexes.} Combinatorial complexes provide the most flexible framework for protein modeling by unifying the hierarchical organization of simplicial/cellular complexes with the set-type relationships of hypergraphs, while removing strict boundary constraints. It is a triple $(S,\mathcal{X},\mathrm{rk})$ where $S$ is a finite vertex set, $\mathcal{X}\subseteq\mathcal{P}(S)\setminus\{\emptyset\}$ is a collection of \textbf{cells}, and $\mathrm{rk}:\mathcal{X}\to\mathbb{Z}_{\ge0}$ is an \textbf{order-preserving} rank function such that $x\subseteq y \Longrightarrow \mathrm{rk}(x)\le\mathrm{rk}(y)$, and all singletons $\{s\}$ for $s\in S$ belong to $\mathcal{X}$ with $\mathrm{rk}(\{s\}) = 0$.

\begin{figure}[hbtp]
\centering
\includegraphics[width=0.9\textwidth]{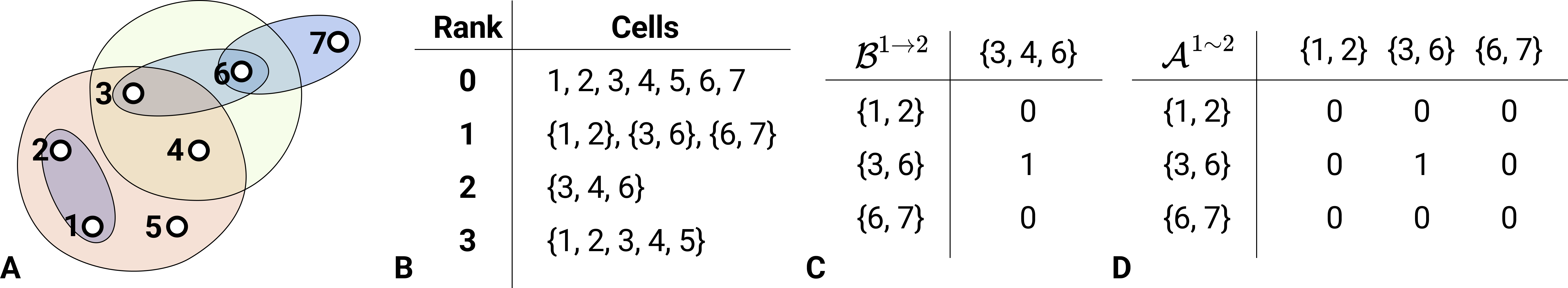}
\caption{\textbf{An example of combinatorial complex.} The visualization of this combinatorial complex (A), its constituent cells organized by rank (B), the incidence matrix showing relationships between rank 1 and rank 2 cells (C), and the adjacent matrix showing relationships between rank 1 via the intermediate rank 2 cells (D).}
\label{fig:ccc}
\end{figure}

As illustrated in Figure~\ref{fig:ccc}, combinatorial complexes provide a flexible representation where cells of different ranks can coexist without strict boundary requirements. The example demonstrates how vertices (rank 0) can participate in edges (rank 1) and higher-order structures (rank 2) simultaneously, with the incidence matrix capturing these relationships mathematically. This flexibility makes combinatorial complexes ideally suited for protein modeling: secondary structures spanning non-contiguous residues can be represented as single 2-cells without artificial boundary edges, the framework naturally accommodates the protein hierarchy from residues to interactions to secondary structures to complete proteins, and diverse geometric arrangements from linear $\alpha$-helices to planar $\beta$-sheets can be accommodated without rigid constraints.

\subsection{Topological Neural Networks}\label{subsec:tnn}
\begin{figure}[hbtp]
    \centering
    \includegraphics[width=0.6\textwidth]{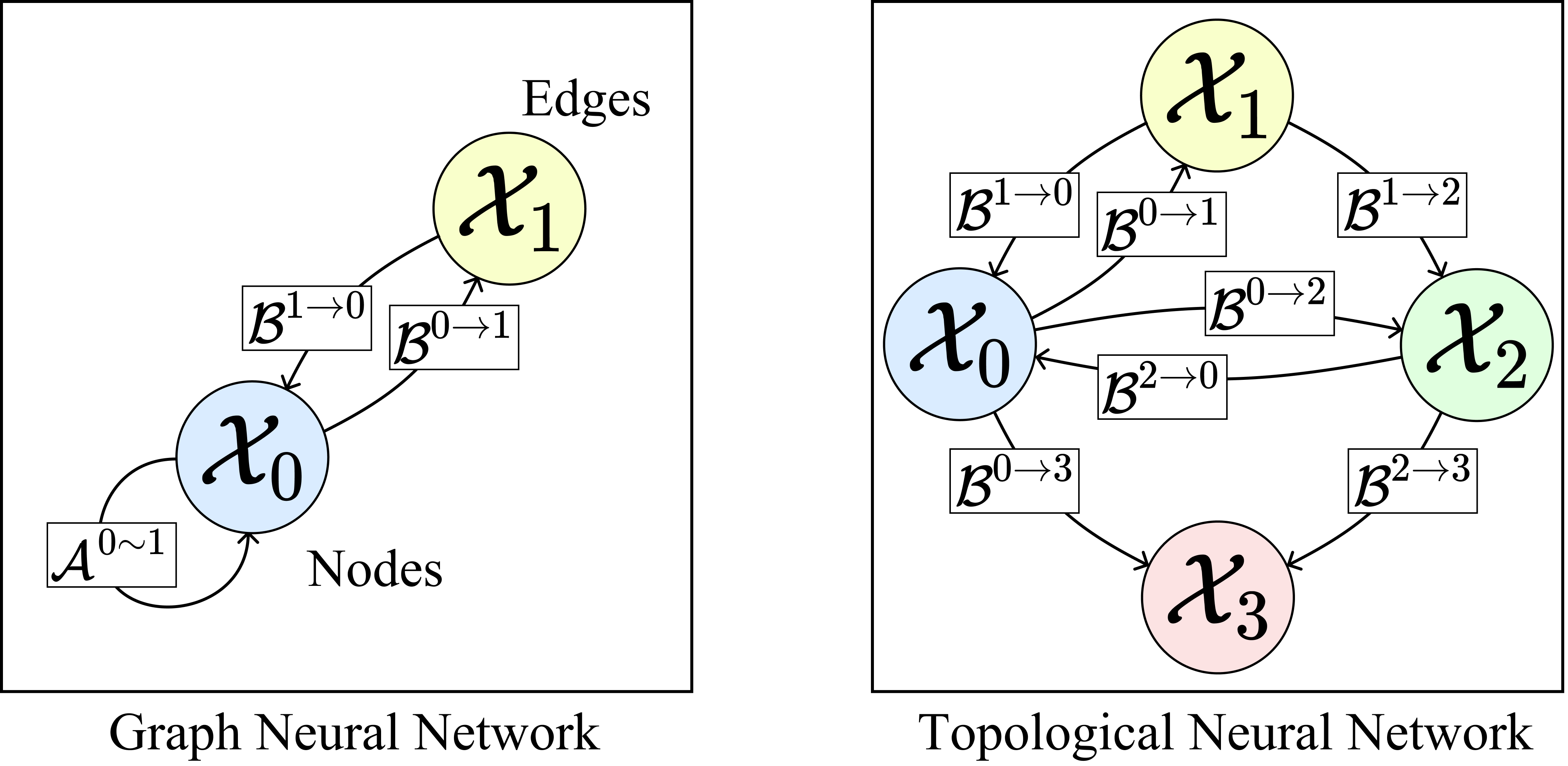}
    \caption{\textbf{Topological Neural Networks vs Graph Neural Networks.} TNNs generalize GNNs by enabling information flow between any ranks in a topological domain. While GNNs are limited to message passing between nodes (rank 0) and edges (rank 1), TNNs allow flexible multi-rank communication: cells can exchange information within the same rank, between different ranks, and across multiple hierarchical levels simultaneously.}
    \label{fig:tnn}
\end{figure}

Topological Neural Networks (TNNs) represent a natural and powerful extension of Graph Neural Networks (GNNs) that transcends the traditional limitations of pairwise node interactions. As illustrated in Figure~\ref{fig:tnn}, while GNNs operate exclusively on nodes and edges (ranks 0 and 1), TNNs generalize the message-passing paradigm to operate on the full spectrum of topological structures within combinatorial complexes, enabling sophisticated multi-rank communication patterns that capture the hierarchical nature of complex systems. To understand this generalization, consider how standard GNNs update node features by aggregating messages from neighboring nodes through the familiar update rule $\mathbf{h}_i^{l+1} = \phi\left(\mathbf{h}_i^{l}, \bigoplus_{j \in \mathcal{A}_i} \psi(\mathbf{h}_i^{l}, \mathbf{h}_j^{l}) \right)$, where $\mathcal{A}_i$ represents the adjacency set of node $i$, $\phi$ is an update function, $\psi$ is a message function, and $\bigoplus$ denotes aggregation. TNNs fundamentally extend this framework to operate on arbitrary topological structures through the generalized update rule:
\begin{align}
    \mathbf{h}_i^{l+1} = \phi \left(\mathbf{h}_i^l, \bigotimes_{\mathcal{N} \in \mathfrak{N}} \bigoplus_{j \in \mathcal{N}_i} \psi_{\mathcal{N}}\left(\mathbf{h}_i^l, \mathbf{h}_j^l\right)\right)
\end{align}
where $i$ represents any cell in the combinatorial complex regardless of rank, $\mathfrak{N}$ is a collection of different neighborhood types, $\bigoplus$ performs intra-neighborhood aggregation, $\bigotimes$ performs inter-neighborhood aggregation, and $\psi_{\mathcal{N}}$ are neighborhood-specific message functions~\parencite{battiloroEquivariantTopologicalNeural2024}. This formulation enables TNNs to simultaneously process information flows within ranks (e.g., between secondary structures), across ranks (e.g., from residues to secondary structures), and through complex multi-hop pathways that traverse multiple hierarchical levels, providing far richer representational capacity than traditional graph-based approaches.

The key innovation of TNNs lies in their sophisticated neighborhood definitions that enable flexible information routing through topological structures via two complementary mechanisms: incident-type and adjacent-type neighborhoods. Incident-type neighborhoods capture direct structural relationships between cells of different ranks through incidence matrices $\mathcal{B}^{r\to r'} \in \{0,1\}^{|N_r| \times |N_{r'}|}$ where $r \neq r'$, encoding which lower-rank cells are constituents of higher-rank cells (when $r < r'$) or which higher-rank cells contain lower-rank cells (when $r > r'$). Specifically, the entry $\mathcal{B}^{r\to r'}[i, j] = 1$ indicates that cell $i$ of rank $r$ and cell $j$ of rank $r'$ have a direct containment relationship, enabling cross-rank message passing that propagates information between hierarchical levels. Adjacent-type neighborhoods capture indirect relationships between cells of the same rank through shared connections to intermediate cells of different ranks, formalized through adjacency matrices $\mathcal{A}^{r\sim r'} \in \mathbb{N}^{|N_r| \times |N_r|}$ where $r \neq r'$. Here, $\mathcal{A}^{r\sim r'}[i, j] = n$ indicates that cells $i$ and $j$ of rank $r$ share connections to $n$ common cells of rank $r'$, enabling within-rank message passing that leverages higher-order structural context. These neighborhoods enable TNNs to process information both within and across hierarchical levels.

To understand how these matrices enable information flow, consider the example in Figure~\ref{fig:ccc}. The incidence matrix shown between rank 1 and rank 2 cells exemplifies $\mathcal{B}^{1\to 2}$, where each entry indicates whether a rank 1 cell (edge) is incident to a rank 2 cell (higher-order structure). For instance, if the matrix shows $\mathcal{B}^{1\to 2}[\{1,2\},\{3, 4, 6\}] = 1$, this means edge $\{1,2\}$ is incident to the rank 2 cell $\{3, 4, 6\}$, enabling direct cross-rank message passing. Similarly, the adjacency matrix $\mathcal{A}^{1\sim 2}$ shown in panel (D) captures relationships between rank 1 cells that share incident rank 2 cells, enabling within-rank message passing through higher-order structures. These neighborhood matrices form the mathematical foundation for flexible topological message passing, defining how information flows both across and within ranks throughout the complex.

\section{Related Work}\label{sec:related-work}

\paragraph{Geometric Graph Neural Networks.}
Geometric Graph Neural Networks (GGNNs) extend standard message-passing networks by incorporating vector features and 3D geometric information while preserving physical symmetries~\parencite{duvalHitchhikersGuideGeometric2024, hanSurveyGeometricGraph2025}. These models capture both graph topology and spatial geometry for structured 3D data like proteins~\parencite{bronsteinGeometricDeepLearning2021a, thomasTensorFieldNetworks2018}. \emph{Invariant GGNNs} operate on transformation-invariant quantities: SchNet~\parencite{schuttSchNetDeepLearning2018} uses continuous-filter convolutions while DimeNet++~\parencite{gasteigerFastUncertaintyawareDirectional2022} incorporates angular information, though they lose orientation-dependent features~\parencite{joshiExpressivePowerGeometric2023}. \emph{Equivariant GGNNs} preserve full Euclidean symmetry through vector features: EGNN~\parencite{satorrasEquivariantGraphNeural2021} supports relative displacements, GVP-GNN~\parencite{jingLearningProteinStructure2020} allows any vector features such as orientations, and GCPNet~\parencite{moreheadGeometrycompletePerceptronNetworks2024} employ local reference frames for non-degenerate learning of vector features.

\paragraph{Protein Representation Learning.}
PRL approaches fall into three main categories: structure-based, sequence-based, and hybrid models. \emph{Structure-based models} represent proteins as residue-level graphs through backbone coordinates and distance features. ProteinMPNN~\parencite{dauparasRobustDeepLearning2022} pioneered coarse-grained message-passing for inverse folding. On top of it, geometric graph models like GearNet~\parencite{zhangProteinRepresentationLearning2022} and GCPNet~\parencite{moreheadGeometrycompletePerceptronNetworks2024} preserve geometric symmetries. Comprehensive evaluation by \textcite{jamasbEvaluatingRepresentationLearning2024} demonstrates their effectiveness across multiple downstream tasks using only 10--20M parameters. \emph{Sequence-based models} like ESM2~\parencite{linEvolutionaryscalePredictionAtomiclevel2023} learns evolutionary patterns from amino acid sequences, while \emph{hybrid models} like SaProt~\parencite{suSaProtProteinLanguage2023} and ESM3~\parencite{hayesSimulating500Million2025} incorporate structural tokens, requiring larger computational resources but serving as powerful foundation models. Despite their success, both types of approach lack inductive bias for hierarchical protein organization from secondary structures to complete folds, motivating our topological approach.

\paragraph{Topological Deep Learning for Molecules.}
While topology-driven architectures remain rare in PRL, advances in small-molecule modeling demonstrate their potential. Cellular-complex approaches like CWN~\parencite{bodnarWeisfeilerLehmanGo2021} and CIN++~\parencite{giustiTopologicalMessagePassing2024} model molecular rings as higher-rank cells, though they operate only on scalar features. Hypergraph-based models like SE3Set~\parencite{wuSE3SetHarnessingEquivariant2024} combine topological structure with geometric equivariance. ETNN~\parencite{battiloroEquivariantTopologicalNeural2024} extends E(n)-equivariant GNNs to combinatorial complexes~\parencite{hajijTopologicalDeepLearning2022}, achieving competitive results on molecular benchmarks and validating the integration of topological and geometric approaches.
However, these methods have limitations when applied to proteins: they focus on small molecules with simple topological features, lack protein-specific architectural considerations, and have not been validated on the complex hierarchical structures characteristic of proteins. Our work addresses these gaps by developing topological neural networks specifically designed for PRL.

\section{Methodology}\label{sec:methodology}

\subsection{Protein Combinatorial Complex}\label{subsec:protein-combinatorial-complex}

\begin{figure}[htbp]
    \centering
    \includegraphics[width=0.75\textwidth]{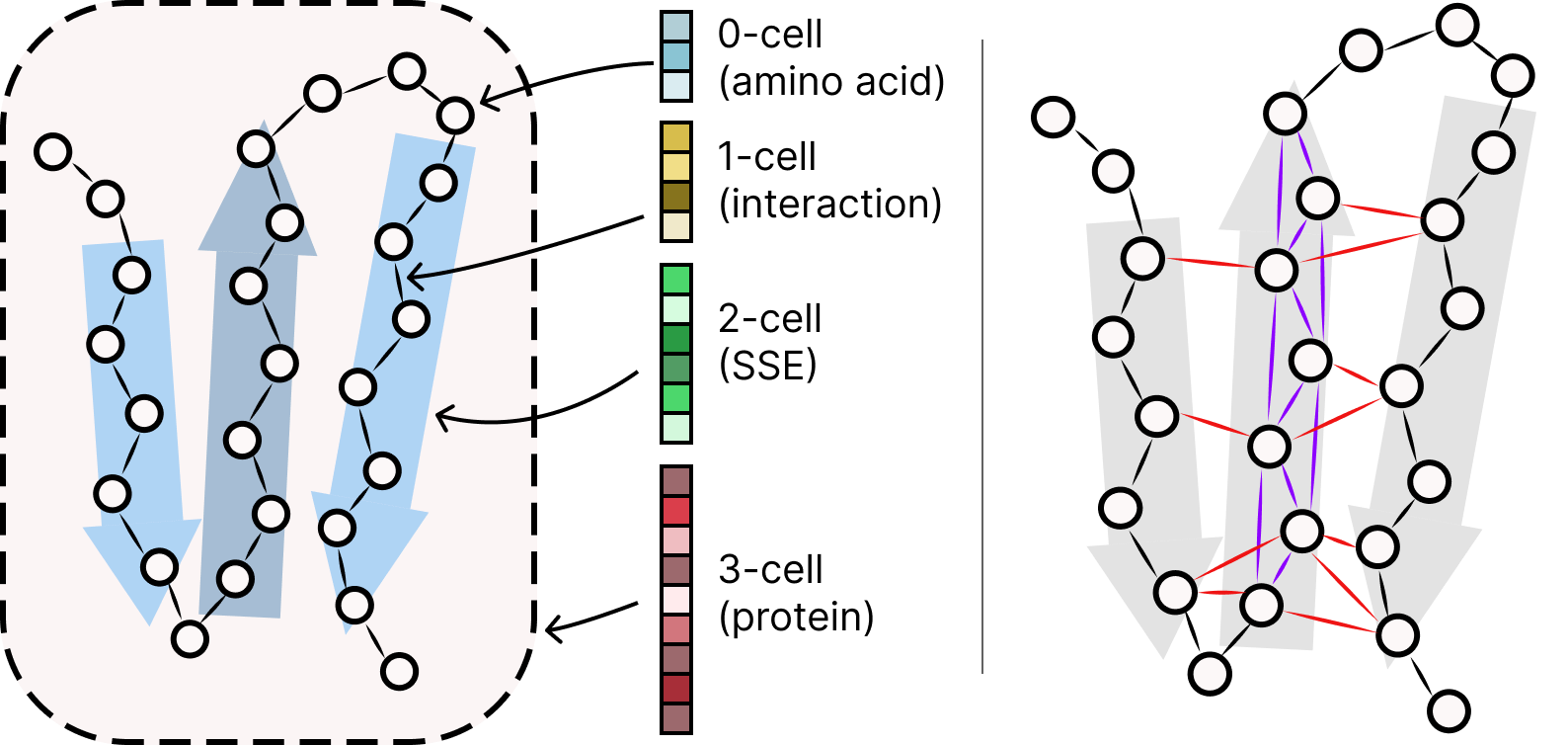}
    \caption{\textbf{Protein Combinatorial Complex.} Left: A PCC example showing amino acid nodes (black circles, 0-cell), interaction edges (black lines, 1-cells), secondary structures (blue arrows, 2-cells), and the protein itself (dashed box, 3-cell). Right: Inner edges (purple) directly incident to 2-cells and outer edges (red) enabling SSE-level message passing via outer-edge neighborhoods.}
    \label{fig:pr_ccc}
\end{figure}

We define a protein combinatorial complex (PCC) as a combinatorial complex $\mathcal{C}=(\mathit{S}, \mathcal{X}, \text{rk})$ where the vertex set $\mathit{S}$ contains all the residues in this protein, the cell set $\mathcal{X}$ contains sets of residues, and the $\text{rk}$ function maps cells from $\mathcal{X}$ to a rank in $R=\{0, 1, 2, 3\}$. As illustrated in Figure~\ref{fig:pr_ccc}, rank 0 to rank 3 cells respectively refer to residues, residue interactions, secondary structure elements (SSEs), and the protein itself. The residue interaction edges are constructed by connecting each residue to its 16 nearest neighbors, and the SSE 2-cells are constructed by including sequentially consecutive 0-cells (residues) that have the same SSE label with a minimum size of three 0-cells.

Notably, unlike the original definition of combinatorial complex where 1-cells are undirected hyperedges~\parencite{hajijTopologicalDeepLearning2022}, PCC's 1-cells are directed pair-wise edges. In this sense, for a directed edge $(i, j)$ pointing from node $i$ to node $j$, only the node $i$ can pass a message to the edge $(i, j)$ via the incidence matrix from rank 0 to rank 1 ($\mathcal{B}^{0 \rightarrow 1}$), and, similarly, edge $(i, j)$ can only pass a message to the node $j$ via $\mathcal{B}^{1 \rightarrow 0}$. By enabling edges to have directions, this design allows calculating SO(3)-equivariant edge frames for scalarizing vector edge features as used in~\parencite{moreheadGeometrycompletePerceptronNetworks2024, duNewPerspectiveBuilding2023}.

Additionally, since each residue can only be assigned to one SSE, none of the 2-cells will have overlap. Consequently, it is impossible for 2-cells in a PCC to communicate with each other directly via down-adjacency, which would only let SSEs propagate tens or even hundreds of the same messages back to themselves. To circumvent this issue and enable 2-cell updates, we design ``outer-edge neighborhoods'' defined as
\begin{align}
    \mathcal{N}^{2 \rightarrow 1}_{outer} &= \mathcal{B}^{2 \rightarrow 0} \cdot \mathcal{B}^{0 \rightarrow 1} - \mathcal{B}^{2 \rightarrow 1} \label{eq:outer_sse2edge} \\
    (\mathcal{N}^{1 \rightarrow 2}_{outer})^\top &= \mathcal{B}^{2 \rightarrow 0} \cdot (\mathcal{B}^{1 \rightarrow 0})^{\top} - \mathcal{B}^{2 \rightarrow 1}, \label{eq:outer_edge2sse}
\end{align}
where $\mathcal{N}^{2 \rightarrow 1}_{outer}$ maps SSEs to edges that originate from within one SSE and terminate in a different SSE, while $(\mathcal{N}^{1 \rightarrow 2}_{outer})^\top$ maps SSEs to edges that terminate within them but originate from a different SSE. These directional neighborhoods enable SSEs to communicate with external cells while avoiding redundant self-connections.

\subsection{Hierarchical Protein Featurization}\label{subsec:hierarchical-protein-featurization}

PCC featurization spans four hierarchical levels with scalar and vector features at each rank. Node and edge features follow established standards~\parencite{jamasbEvaluatingRepresentationLearning2024, tanProteinRepresentationLearning2024}, while we introduce novel SSE and protein-level features capturing multi-scale geometric and contextual information. See Appendix~\ref{sec:featurization-details} for a complete list of features used.

\textbf{Nodes and Edges.} We use $C_\alpha$ backbone features. Node scalar features include 23-dimensional amino acid one-hot encoding, 21-dimensional 3Di~\parencite{vankempenFastAccurateProtein2024} one-hot encoding, 16-dimensional positional encoding, virtual bond and torsion angles $\alpha$ and $\kappa$, and backbone dihedral angles $\phi$, $\psi$, and $\omega$ (sine/cosine encoded). Vector features are displacement vectors to neighboring residues and tetrahedral geometry~\parencite{tanProteinRepresentationLearning2024}. Edge features include Euclidean distance, and positional encoding of that distance, and displacement vectors between connected nodes.

\textbf{Secondary Structures.} SSE features capture geometric and contextual properties. Scalar features include 3-dimensional SSE type encoding (helix, strand, coil from DSSP~\parencite{kabschDictionaryProteinSecondary1983a}), SSE size, 20-dimensional positional encoding for start/end residues, consecutive angles between neighboring SSEs, and five shape descriptors derived from eigenvalues (linearity, planarity, scattering, omnivariance, anisotropy). Vector features include six displacement vectors between COM/start/end/midpoint positions, three principal eigenvectors disambiguated by following~\textcite{broResolvingSignAmbiguity2008}, and displacement vectors to adjacent SSEs and protein COM.

\textbf{Proteins.} Global protein features capture sequence composition, SSE distribution, and 3D geometry. Scalar features include protein size, amino acid composition frequencies, SSE type frequencies with size statistics, eight shape descriptors from eigenvalues, radius of gyration, and contact density measures. Vector features include three disambiguated global eigenvectors as canonical frames and displacement vectors to the ten farthest/nearest residues from protein COM.

\subsection{Topology-Complete Perceptron}

\begin{figure}[htbp]
    \centering
    \includegraphics[width=0.75\textwidth]{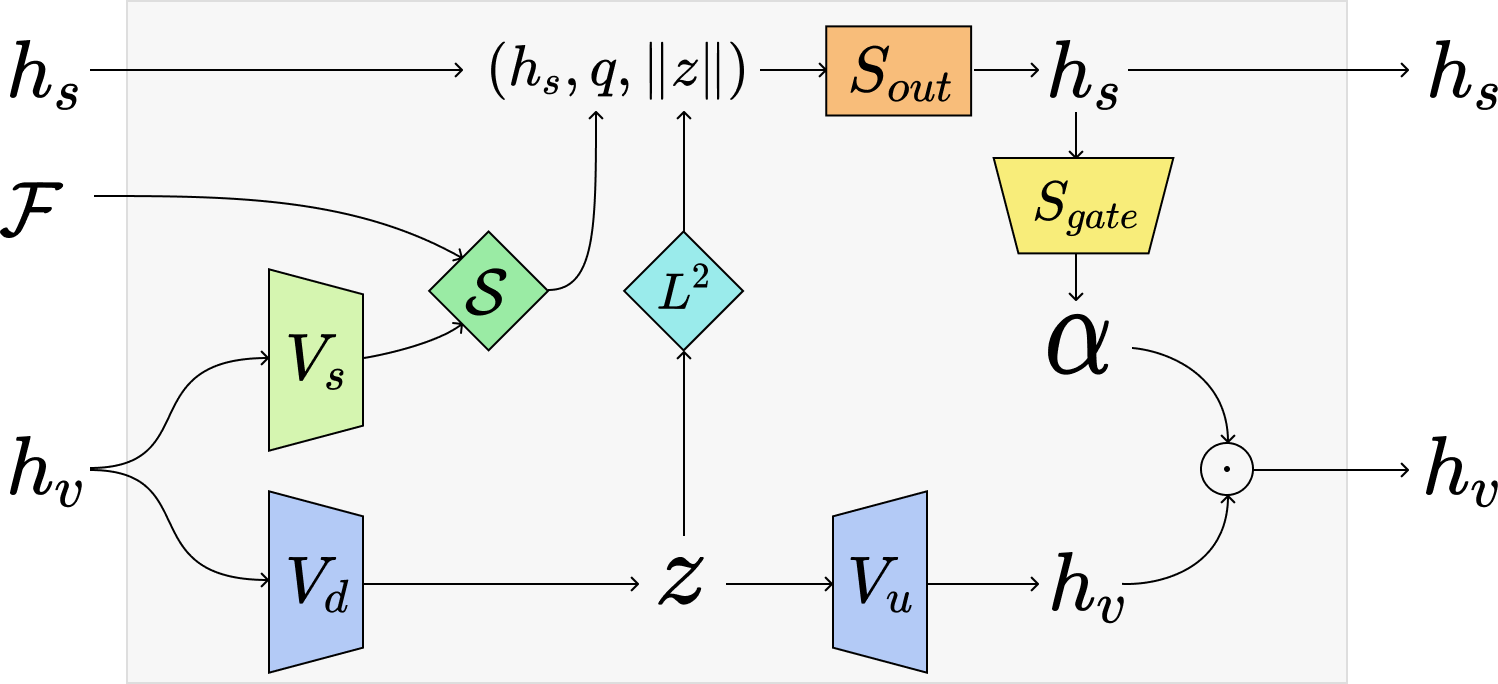}
    \caption{\textbf{Architecture of Topology-Complete Perceptron (TCP) module.} The module processes scalar features $\mathbf{h}_s$, vector features $\mathbf{h}_v$, and rank-specific frames $\mathcal{F}^{(r)}_i$ through dual pathways: vector features are reduced via MLPs ($V_s$, $V_d$), scalarized using localized frames, concatenated with scalar features, and processed by output MLPs ($S_{out}$, $V_u$). The final vector output is gated by scalar features through $S_{gate}$ to maintain geometric structure while enabling expressive computations.}
    \label{fig:tcp}
\end{figure}

The Topology-Complete Perceptron (TCP) module is a TDL generalization of the Geometry-Complete Perceptron (GCP)~\parencite{moreheadGeometrycompletePerceptronNetworks2024}. Unlike GCP, which operates only on node and edge features, TCP processes information from cells of arbitrary topological rank within our PCC structure. The module accepts scalar features $\mathbf{h}_s \in \mathbb{R}^{d_s}$, vector features $\mathbf{h}_v \in \mathbb{R}^{d_v \times 3}$, and rank-specific localized frames $\mathcal{F}^{(r)}_i \in \mathbb{R}^{3 \times 3}$, processing them through dual pathways that preserve geometric structure while enabling expressive scalar computations.

The module begins by reducing vector feature dimensions using MLPs $V_s$ and $V_d$:
\begin{align}
s &= \sigma(V_s(\mathbf{h}_v)) \in \mathbb{R}^{3 \times 3} \\
z &= \sigma(V_d(\mathbf{h}_v)) \in \mathbb{R}^{\frac{d_v}{\lambda} \times 3},
\end{align}
where $\sigma$ represents nonlinear activation, $d_v$ is the vector dimension, and $\lambda$ is the bottleneck parameter. The reduced features are then integrated with scalar features through scalarization and normalization:
\begin{align}
\mathbf{h}_s^\prime = (\mathbf{h}_s, \mathcal{S}^{(r)}_i(s), \|z\|_2) \in \mathbb{R}^{d_s + 9 + \frac{d_v}{\lambda}}.
\end{align}
Final outputs are computed through dedicated MLPs with scalar gating:
\begin{align}
\mathbf{h}_{s,out} &= \sigma(S_{out}(\mathbf{h}_s^\prime)) \in \mathbb{R}^{d_s} \\
\mathbf{h}_v^{\prime} &= \sigma(V_u(z)) \in \mathbb{R}^{d_v \times 3} \\
\mathbf{h}_{v,out} &= \mathbf{h}_v^{\prime} \odot \sigma_g(S_{gate}(\mathbf{h}_{s,out})) \in \mathbb{R}^{d_v \times 3},
\end{align}
where $\odot$ denotes element-wise multiplication and $\sigma_g$ is sigmoid activation.

The key innovation of TCP lies in its scalarization capability for SSE and protein-level vector features, which is crucial for maintaining SE(3)-equivariance across all topological ranks. SE(3)-equivariance ensures that protein representations remain equivariant to global rotations and translations while preserving sensitivity to reflection---a fundamental requirement since protein function depends on 3D structure but not on global orientation in space. While the computational architecture remains similar to GCP~\parencite{moreheadGeometrycompletePerceptronNetworks2024}, the scalarization operation $\mathcal{S}^{(r)}_i(\cdot)$ enables TCP to process geometric information from arbitrary-rank cells through localized coordinate frames, allowing unified treatment of residues, interactions, secondary structures, and protein-level features within a single framework.

Our approach employs edge-centric scalarization for enhanced geometric sensitivity. Rather than using cell-specific frames, we project vector features onto frames of associated edges and aggregate the results. Edge-level frames form the foundation of our geometric representation, utilizing source and target node positions:
\begin{align}
    \mathcal{F}^{(1)}_{(i, j)} &= \left(
     \frac{\mathbf{x}_j - \mathbf{x}_i}{\|\mathbf{x}_j - \mathbf{x}_i\|},
     \frac{\mathbf{x}_j \times \mathbf{x}_i}{\|\mathbf{x}_j \times \mathbf{x}_i\|},
     \frac{\mathbf{x}_j - \mathbf{x}_i}{\|\mathbf{x}_j - \mathbf{x}_i\|} \times \frac{\mathbf{x}_j \times \mathbf{x}_i}{\|\mathbf{x}_j \times \mathbf{x}_i\|}
    \right)\\
    \mathcal{S}^{(1)}_{(i,j)}(\mathbf{h}_{{(i,j)},v}^{(1)}) &= \text{flatten}\left(\mathbf{h}_{{(i,j)},v}^{(1)} \cdot \mathcal{F}^{(1)}_{(i, j)}\right).
\end{align}
For nodes (rank 0), vector features are projected onto frames of incident edges:
\begin{align}
\mathcal{S}^{(0)}_i(\mathbf{h}_{i,v}^{(0)}) = \text{flatten}\left(\frac{1}{|\mathcal{B}^{0 \to 1}_i|}\sum_{(i, j) \in \mathcal{B}^{0 \to 1}_i} \mathbf{h}_{i,v}^{(0)} \cdot \mathcal{F}^{(1)}_{(i, j)}\right).
\end{align}
For SSEs (rank 2), we use outer-edge neighborhoods to capture inter-SSE relationships:
\begin{align}
\mathcal{S}^{(2)}_i(\mathbf{h}_{i,v}^{(2)}) = \text{flatten}\left(\frac{1}{|\mathcal{N}^{2 \to 1}_i|}\sum_{(l, j) \in \mathcal{N}^{2 \to 1}_i} \mathbf{h}_{i,v}^{(2)} \cdot \mathcal{F}^{(1)}_{(l, j)}\right),
\end{align}
where $\mathcal{N}^{2 \to 1} = \mathcal{N}^{2 \rightarrow 1}_{outer} \cap (\mathcal{N}^{1 \rightarrow 2}_{outer})^\top$ represents the intersection of the outer-edge neighborhoods.

Protein-level frames employ principal component analysis for global orientation, where edge-centric construction is not meaningful due to the lack of meaningful inter-protein edges $\mathcal{F}^{(3)}_{i} = (\mathbf{v}^i_1, \mathbf{v}^i_2, \mathbf{v}^i_3)$
The principal component eigenvectors $\mathbf{v}^i_{1,2,3}$ are disambiguated to ensure consistent frame orientation across different protein conformations. We employ the disambiguation procedure proposed by \textcite{broResolvingSignAmbiguity2008}, using the farthest residue from protein center as an anchor vector to resolve sign ambiguities and maintain geometric consistency.

While individual rank-specific frames were also developed in our work (see Appendix~\ref{sec:alternative-frames}), empirical results showed that the edge-centric approach provides superior geometric expressiveness by implicitly encoding biologically relevant bonding patterns.
\subsection{Topology-Complete Perceptron Network}

\begin{figure}[htbp]
    \centering
    \includegraphics[width=\textwidth]{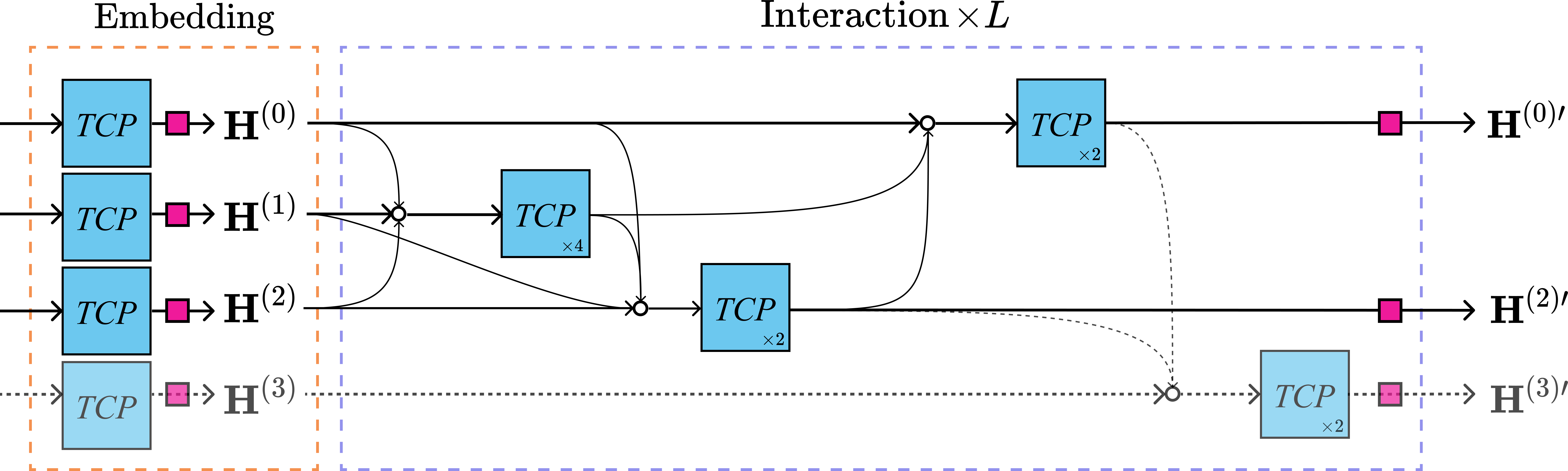}
    \caption{\textbf{Architecture of the Topology-Complete Perceptron Network (TCPNet).} The network processes both scalar and vector features through parallel pathways from rank 0 to 3. The network starts with embedding the raw features of each rank with separate TCP modules. Then $L$ interaction layers are applied for message passing. The protein (rank 3) channel is optional and can be replaced by applying a standard pooling function to the final node embedding. The pink boxes represent GVP norms~\parencite{jingLearningProteinStructure2020}, and the white circles denote concatenation. }
    \label{fig:tcpnet_model}
\end{figure}

TCPNet represents our complete architecture for hierarchical protein structure analysis, seamlessly integrating scalar and vector features across all topological ranks within the PCC structure. As illustrated in Figure~\ref{fig:tcpnet_model}, the architecture comprises an embedding module for mapping raw features to a latent space followed by $L$ interaction modules for inter- and intra-rank message passing.

\paragraph{Embedding Module.} The embedding phase transforms heterogeneous raw features into unified geometric representations through rank-specific processing. First, GVP layer normalization~\parencite{jingLearningProteinStructure2020} is applied to stabilize vector features:
\begin{align}
    \text{LN}(\mathbf{H}) = \left(\text{LN}_s(\mathbf{H}_s), \frac{\mathbf{H}_v}{\sqrt{\|\mathbf{H}_v\|^2_2 / |\mathbf{H}|}}\right).
\end{align}
Then, rank-specific TCP embedding modules transform the normalized features:
\begin{align}
    \{\mathbf{H}^{(r)}\}_{r \in R} = \left\{\varphi_{\text{emb}}^{(r)}\left(\text{LN}\left(\mathbf{H}^{(r)}_0\right)\right)\right\}_{r \in R},
\end{align}
where $\varphi_{\text{emb}}^{(r)}$ represents rank-$r$ specific TCP embedding modules that map heterogeneous raw features to fixed-dimensional geometric representations.

\paragraph{Hierarchical Message Passing.} The interaction component employs a four-step hierarchical message passing scheme that systematically propagates information across all topological ranks in our PCC structure. This coordinated flow enables comprehensive structural reasoning by allowing residues, interactions, secondary structures, and the global protein to exchange information in a biologically meaningful order. The message passing follows a carefully designed sequence: first computing edge-level messages that integrate local residue and SSE information, then updating SSE representations using these enriched edge messages, followed by refining residue features with hierarchical context, and finally creating global protein representations. This bottom-up-down approach ensures that each rank receives contextually relevant information from both local and global perspectives.

\emph{Step 1: Edge-level message computation.} We begin by computing messages at interaction edges, which serve as communication channels between residues. Each edge message aggregates information from its source and target residues, the edge's own features, and the secondary structures containing these residues:
\begin{align}
    \mathbf{m}_{ij} = \phi^{(1)}\left(\mathbf{h}^{(0)}_i, \mathbf{h}^{(0)}_j, \mathbf{h}^{(1)}_{ij}, \mathbf{n}_i, \mathbf{n}_j\right),
\end{align}
where SSE features are mapped to each residue $i$ according to structural membership:
\begin{align}
    \mathbf{n}_i = \begin{cases}
                       \mathbf{h}^{(2)}_k & \text{if } \exists k \in \mathcal{X}, \text{rk}(k) = 2 \text{ and } i \in k \\
                       \mathbf{0} & \text{otherwise}
    \end{cases}.
\end{align}
The message computation incorporates TCP residual connections and scalar attention to focus on the most relevant interactions:
\begin{align}
    \mathbf{h}^{l+1}_{ij} &= \varphi_{\text{msg}}^l(\mathbf{h}^l_{ij}) + \mathbf{h}^l_{ij} \\
    \mathbf{m}_{ij} &= (\mathbf{h}_{ij,s}^{\text{fin}} \odot \sigma_{\text{att}}(\mathbf{w}^\top \cdot \mathbf{h}_{ij,s}^{\text{fin}}), \mathbf{h}_{ij,v}^{\text{fin}}).
\end{align}

\emph{Step 2: SSE-level information integration.} Having computed enriched edge messages, we now update secondary structure representations by aggregating information from multiple sources. Each SSE collects information from its constituent residues, internal edges, and crucially, external connections through our outer-edge neighborhoods that enable inter-SSE communication:
\begin{align}
    \mathbf{u}_{i}^{(2)} = \phi^{(2)}\left(
         \mathbf{h}^{(2)}_i, \bigoplus_{j \in \mathcal{B}^{2 \to 0}_i} \mathbf{h}^{(0)}_j,
         \bigoplus_{j \in \mathcal{B}^{2 \to 1}_i} \mathbf{h}^{(1)}_{j},
         \bigoplus_{(j, k) \in \mathcal{N}^{2 \to 1}_i} \mathbf{m}_{jk}
    \right).
\end{align}
This aggregation captures both the internal structure of each SSE and its relationship to neighboring secondary structures, providing a comprehensive view of local protein architecture.

\emph{Step 3: Residue-level refinement.} With updated SSE representations, we refine residue features by incorporating hierarchical context from both their parent secondary structures and neighboring interactions. This step enables residues to benefit from both local edge information and broader structural context:
\begin{align}
    \mathbf{u}_{i}^{(0)} = \phi^{(0)}\left(
                                         \mathbf{h}^{(0)}_i,
                                         \mathbf{m}_{i}^{(2)},
                                         \bigoplus_{j \in (\mathcal{B}^{1 \to 0}_i)^\top} \mathbf{m}_{ji}
    \right),
\end{align}
where updated SSE information is propagated to constituent residues:
\begin{align}
    \mathbf{m}_i^{(2)} = \begin{cases}
                             \mathbf{u}^{(2)}_k & \text{if } \exists k \in \mathcal{X}, \text{rk}(k) = 2 \text{ and } i \in k \\
                             \mathbf{0} & \text{otherwise}
    \end{cases}.
\end{align}

\emph{Step 4: Global protein representation.} Finally, we create protein-level representations by aggregating information across all hierarchical levels. This global view captures the overall protein architecture while maintaining sensitivity to local structural details:
\begin{align}
    \mathbf{u}_{i}^{(3)} = \phi^{(3)}\left(\mathbf{u}^{(0)}_i, \mathbf{u}_{i}^{(2)}, \mathbf{h}_{i}^{(3)}\right).
\end{align}

\emph{Feature updates.} All rank-specific representations are updated using residual connections and layer normalization to ensure training stability and gradient flow:
\begin{align}
    \mathbf{h}_{i}^{(r)'} = \text{LN}(\mathbf{u}_{i}^{(r)} + \mathbf{h}_{i}^{(r)}), r \in \{0,2,3\}.
\end{align}

This systematic four-step message passing scheme enables TCPNet to capture complex structural dependencies and geometric relationships while maintaining computational efficiency. The hierarchical flow ensures that information propagates meaningfully across biological scales, from local residue interactions to global protein architecture.

\paragraph{Readout.} We implement two distinct readout strategies for outputting graph-level embeddings. First, pooling node embeddings by summing or averaging them like conventional GNNs. In this case, the protein-level message passing is not used. Second, we can directly use the protein embedding as the graph representation. The first approach requires node embeddings to capture both local and global information, while the second approach allows for clearer separation between node-level and graph-level representations, offering greater capacity but with increased risk of overfitting.

\section{Experiments}
\subsection{Experimental Setup}

We evaluate our approach on four protein structure analysis tasks at different structural levels: (1) node-level inverse folding (23-class amino acid prediction) and three graph-level tasks including (2) fold classification, (3) cellular component prediction (from the Gene Ontology task), and (4) antibody developability prediction.

\textbf{Datasets.} Inverse folding uses CATH4.4~\parencite{knudsenTheCATHDatabase2010, wamanCATHV44Major2025} (32,652/1,000/1,000 train/val/test, 40\% similarity filtering). Fold classification uses SCOP 1.75~\parencite{murzinSCOPStructuralClassification1995} with three difficulty splits (Family/Superfamily/Fold: 1,272/1,254/718 test proteins). Cellular component prediction uses the Gene Ontology dataset~\parencite{gligorijevicStructurebasedProteinFunction2021a} with 16,019/1,294/1,714 train/val/test splits. Antibody developability uses SabDab~\parencite{dunbarSAbDabStructuralAntibody2014} with 1,669/241/478 train/val/test splits.

\textbf{Configuration.} All models use 6-layer architectures with 128-dimensional scalars and 16-dimensional vectors. Training uses Adam optimization (lr=0.001), ReduceOnPlateau scheduling, and early stopping on A100 GPUs. We compare TCPNet, GVP-TNN, and ETNN~\parencite{battiloroEquivariantTopologicalNeural2024} against GCPNet~\parencite{moreheadGeometrycompletePerceptronNetworks2024}, GVP-GNN~\parencite{jingLearningProteinStructure2020}, and EGNN~\parencite{satorrasEquivariantGraphNeural2021} from ProteinWorkshop~\parencite{jamasbEvaluatingRepresentationLearning2024}. Extended experimental details including datasets and model configurations are provided in the Appendix~\ref{app:details}.

\subsection{Comparing TCPNet with GGNNs}

Table~\ref{tab:performance-tcp-vs-ggnns} presents a comprehensive comparison of TCPNet against three state-of-the-art GGNN architectures across multiple protein structure analysis tasks. Overall, TCPNet demonstrates strong performance, achieving best results in fold classification across all splits and in antibody developability prediction. While GVP-GNN leads in inverse folding with lower perplexity and higher accuracy, TCPNet consistently outperforms its direct GGNN counterpart GCPNet across all tasks. Most notably, TCPNet shows remarkable stability when using only structural features, maintaining competitive performance even without sequence information in the cellular component prediction and antibody developability tasks where other models see significant degradation. These results validate TCPNet's enhanced ability to capture and leverage structural information through its topological approach.

\begin{table}[ht]
    \centering
    \caption{\textbf{Model performance comparison across protein structure analysis tasks.} Each task-specific metric represents performance on held-out test sets, with \textbf{bold} indicating best performance and \underline{underlined} values showing second best. For cellular component prediction and antibody developability, results are shown as structure-only/structure+sequence, while other tasks use structural features only.}
    \label{tab:performance-tcp-vs-ggnns}
    \resizebox{\textwidth}{!}{%
        \begin{tabular}{lccccccc}
            \toprule
            & \multicolumn{2}{c}{Inverse Folding}
            & \multicolumn{3}{c}{Fold Classification}
            & \multicolumn{1}{c}{Cellular Component}
            & \multicolumn{1}{c}{Antibody Dev.} \\
            \cmidrule(lr){2-3}\cmidrule(lr){4-6}\cmidrule(lr){7-7}\cmidrule(lr){8-8}
            Model
            & Perplexity & Accuracy
            & Fold Acc.\ & Superfam.\ Acc.\ & Fam.\ Acc.
            & F1$_{\max}$   & AUPRC \\
            \midrule
            TCPNet
            & \underline{5.822}    & \underline{0.441}
            & \textbf{0.433}   & \textbf{0.558}     & \textbf{0.971}
            & \textbf{0.392}/0.398    & \textbf{0.854}/\underline{0.874} \\
            GCPNet
            & 6.187 & 0.427
            & 0.384            & 0.509              & 0.954
            & \underline{0.388}/\textbf{0.408} & 0.769/\textbf{0.878} \\
            GVP-GNN
            & \textbf{5.280}                & \textbf{0.474}
            & 0.380            & \underline{0.555}   & \underline{0.968}
            & 0.386/\underline{0.403}             & \underline{0.840}/0.818 \\
            EGNN
            & 7.858                & 0.347
            & \underline{0.401}           & 0.552              & 0.967
            & 0.360/0.387             & 0.814/0.781 \\
            \bottomrule
        \end{tabular}%
    }
\end{table}

A detailed analysis of these results reveals several key insights about TCPNet's performance characteristics and the effectiveness of topological enhancements across different task domains:

\textbf{Fold Classification Excellence.} TCPNet achieves best performance across all fold classification splits, with a notable 3\% improvement on the challenging Fold split compared to the second-best model. This task represents the most favorable domain for topological enhancements, as fold classes are directly defined by secondary structure organization patterns. The 3\% improvement on the hardest fold split demonstrates TCPNet's enhanced ability to generalize learned fold patterns to unseen superfamilies, where training and test sets share minimal structural similarity.

\textbf{Structure-Only Robustness.} TCPNet demonstrates exceptional stability when sequence information is unavailable, experiencing only minimal performance decline (0.6\%/2\% on cellular component/antibody tasks) compared to GGNNs which suffer up to 10\% degradation. This robustness stems from TCPNet's hierarchical architecture that effectively captures structural context through multi-rank message passing, enabling superior structural representation learning even in scenarios where sequence information may be noisy or unavailable.

\textbf{Competitive Performance Across Tasks.} While TCPNet consistently outperforms its direct counterpart GCPNet across all tasks, achieving second-best results in inverse folding with 6\% perplexity reduction and 1.4\% accuracy improvement, GVP-GNN has a notable lead in inverse folding (3.3\% accuracy advantage). This suggests that topological enhancements provide measurable benefits but may be constrained by the representational capacity of the underlying GGNN architecture.

\subsection{Impact of TDL Enhancement}

\begin{table}[ht]
    \centering
    \caption{\textbf{Impact of TDL enhancement on model performance across protein structure analysis tasks.} Results compare three GTNN models (TCPNet, GVP-TNN, ETNN) against their GGNN counterparts (GCPNet, GVP-GNN, EGNN). Task metrics are reported on test sets with \underline{underline} highlighting better performance between each GTNN model and its GGNN baseline and \textbf{bold} representing the best among all models. For cellular component prediction and antibody developability, results are shown as structure-only/structure+sequence, while other tasks use structural features only.}

    \label{tab:performance-tdl-enhancement}
    \resizebox{\textwidth}{!}{%
        \begin{tabular}{lccccccc}
            \toprule
            & \multicolumn{2}{c}{Inverse Folding}
            & \multicolumn{3}{c}{Fold Classification}
            & \multicolumn{1}{c}{Cellular Component}
            & \multicolumn{1}{c}{Antibody Dev.} \\
            \cmidrule(lr){2-3}\cmidrule(lr){4-6}\cmidrule(lr){7-7}\cmidrule(lr){8-8}
            Model
            & Perplexity & Accuracy
            & Fold Acc.\ & Superfam.\ Acc.\ & Fam.\ Acc.
            & F1$_{\max}$   & AUPRC \\
            \midrule
            TCPNet
            & \underline{5.822}    & \underline{0.441}
            & \textbf{\underline{0.433}}   & \underline{0.558}    & \textbf{\underline{0.971}}
            & \textbf{\underline{0.392}}/0.397    & \textbf{\underline{0.854}}/0.874 \\
            GCPNet
            & 6.187 & 0.427
            & 0.384            & 0.509              & 0.954
            & 0.388/\textbf{\underline{0.408}} & 0.769/\textbf{\underline{0.878}} \\
            \midrule
            GVP-TNN
            & 5.372               & 0.462
            & \underline{0.431}& \textbf{\underline{0.575}}     & 0.964
            & 0.360/0.343                 & 0.783/\underline{0.836} \\
            GVP-GNN
            & \textbf{\underline{5.280}}                & \textbf{\underline{0.474}}
            & 0.380            & 0.555   & \underline{0.968}
            & \underline{0.386}/\underline{0.403}             & \underline{0.840}/0.818 \\
            \midrule
            ETNN
            & 8.492                & 0.328
            & 0.341            & 0.411              & 0.917
            & \underline{0.384}/0.372                 & 0.800/\underline{0.785} \\
            EGNN
            & \underline{7.858}                & \underline{0.347}
            &  \underline{0.401}           & \underline{0.552}              & \underline{0.967}
            & 0.360/\underline{0.387}             & \underline{0.814}/0.781 \\
            \bottomrule
        \end{tabular}%
    }
\end{table}

We evaluate three TDL-enhanced models against their GGNN counterparts to assess the overall impact of topological enhancement. \textbf{TCPNet} implements comprehensive hierarchical message passing with dedicated reference frames and complete GCPNet redesign. \textbf{GVP-TNN}, developed in this work, adds SSE-level message passing to GVP-GNN while preserving original GVP modules unchanged. \textbf{ETNN} adapts the existing ETNN framework~\parencite{battiloroEquivariantTopologicalNeural2024} to protein tasks by allowing residue nodes to receive rank 2 (SSE) adjacency information without dedicated SSE update mechanisms. Implementation details are provided in the Appendix. Table~\ref{tab:performance-tdl-enhancement} reveals several insights about topological enhancement effectiveness.

First, the effectiveness of TDL enhancement depends critically on how deeply it is integrated into the model architecture. TCPNet shows consistent improvements over GCPNet across all tasks through its comprehensive hierarchical message passing approach. In contrast, GVP-TNN and ETNN show mixed results with their more limited enhancements. ETNN's performance degradation compared to EGNN (8.492 vs 7.858 perplexity in inverse folding) demonstrates that superficial SSE feature addition without dedicated update mechanisms can be counterproductive, as maintaining pairwise residue interactions without updating SSE features creates representational inconsistencies.

The benefits of topological enhancement also vary by task. Fold classification tasks, which inherently depend on SSE organization patterns, show the strongest response. TCPNet and GVP-TNN achieve substantial improvements on the challenging Fold split (5\% accuracy gain), validating that SSE-level message passing effectively captures structural motifs defining fold classes~\parencite{murzinSCOPStructuralClassification1995}. In contrast, inverse folding shows more modest gains, suggesting local residue environments may be adequately captured by traditional geometric features alone.

The generally moderate improvements (typically 1-5\%) can be attributed to two key factors: First, proteins have inherently limited numbers of SSEs compared to residues (typically 3-15 SSEs vs 50-500 residues), constraining their influence on message passing. Second, current TDL models represent incremental modifications rather than ground-up topological designs. TCPNet's success suggests that comprehensive architectural redesign is necessary to fully leverage topological structure in protein representation learning.

\section{Conclusion}\label{sec:conclusion}

We introduce Topotein, a topological deep learning framework that models hierarchical protein structures---from residues to secondary structures---through PCC and TCPNet. Our approach achieves significant performance improvements across four protein analysis tasks, particularly excelling at challenging fold classification where traditional methods struggle. Topotein's design enables integration with existing protein analysis pipelines and could potentially enhance state-of-the-art models like ESM~\parencite{hayesSimulating500Million2025} by providing complementary structural insights. Beyond immediate protein applications, the framework's principled topological approach establishes a new paradigm for modeling hierarchical biological structures, with promising extensions to other complex systems in structural biology and beyond.

\printbibliography

\newpage
\appendix
\section{Notations}\label{sec:notations}
\begin{table}[ht]
    \centering
    \caption{Basic notations and definitions throughout this work.}
    \label{tab:notation}
    \begin{tabular}{@{}p{0.22\textwidth}|p{0.78\textwidth}@{}}
        \toprule
        \textbf{Notation} & \textbf{Description} \\

        \midrule
        \rowcolor{gray!15}\multicolumn{2}{c}{\textbf{Data Structure}} \\
        \midrule
        $\mathcal{G}=(\mathcal{V}, \mathcal{E})$
        & A graph with vertice set $\mathcal{V}$ and edge set $\mathcal{E}$. \\
        $\mathbf{x}_i$ & Position of node $i$, $\mathbf{x}_i\in\mathbb{R}^3$. \\
        $\mathbf{h}_i = (\mathbf{h}_{i,s},\mathbf{h}_{i,v})$
        & Features of node $i$, $\mathbf{h}_i\in\mathbb{R}^{d_{\mathcal{V}}}$. Optionally, $\mathbf{h}_i$ can be split to scalar and vector parts:
        $\mathbf{h}_{i,s}\in\mathbb{R}^{d_{\mathcal{V},s}}$,
        $\mathbf{h}_{i,v}\in\mathbb{R}^{d_{\mathcal{V},v}\times3}$. \\
        $\mathbf{e}_{ij} = (\mathbf{e}_{ij,s},\mathbf{e}_{ij,v})$
        & Features of edge $(i,j)$, $\mathbf{e}_{ij}\in\mathbb{R}^{d_{\mathcal{E}}}$. Optionally, $\mathbf{e}_{ij}$ can be split to scalar and vector parts:
        $\mathbf{e}_{ij,s}\in\mathbb{R}^{d_{\mathcal{E},s}}$,
        $\mathbf{e}_{ij,v}\in\mathbb{R}^{d_{\mathcal{E},v}\times3}$. \\
        $\mathcal{F}_i, \mathcal{F}_{ij}$ & Localized frames of node $i$ and edge $(i,j)$. \\

        \midrule
        $\mathcal{C}=(\mathit{S}, \mathcal{X}, \text{rk})$
        & A combinatorial complex with a vertice (0-cell) set $\mathit{S}$, a cell set $\mathcal{X}$, and a rank function $\text{rk} \colon \mathcal{X} \rightarrow \mathbb{N}$ that maps a cell to its rank. \\
        $\mathbf{h}^{(r)}_i = (\mathbf{h}^{(r)}_{i,s}, \mathbf{h}^{(r)}_{i,v})$
        & Features of the $i$-th cell of rank $r$, $\mathbf{h}^{(r)}_i \in \mathbb{R}^{d_r}$. Optionally, $\mathbf{h}^{(r)}_i$ can be split to scalar and vector parts: $\mathbf{h}^{(r)}_{i,s} \in \mathbb{R}^{d_r,s}$ and $\mathbf{h}^{(r)}_{i,v} \in \mathbb{R}^{d_r,v \times 3}$.  \\
        $\mathcal{F}^{(r)}_i$ & The localized frame of the $i$-th cell of rank $r$. \\
        $\mathcal{M}^{(r)}_i$ & The center of mass of the $i$-th cell of rank $r$. \\
        $\mathcal{N}^{r \rightarrow r'}, \mathcal{N}^{r \sim r'}$ & General neighborhood matrix from rank $r$ to rank $r'$ and neighborhood matrix of rank $r$ via rank $r'$.\\
        $\mathcal{B}^{r \rightarrow r'}$ & Incidence matrix from rank $r$ to rank $r'$.   \\
        $\mathcal{L}^{r \sim r'}, \mathcal{A}^{r \sim r'}, \mathcal{D}^{r \sim r'}$ & Laplacian/adjacency/degree matrix of rank $r$ via rank $r'$. \\ & $\mathcal{L}^{r \sim r'} = \mathcal{B}^{r \rightarrow r'} \cdot \mathcal{B}^{r' \rightarrow r}$, $\mathcal{A}^{r \sim r'} = \mathcal{L}^{r \sim r'} - \mathcal{D}^{r \sim r'}$.       \\

        \midrule
        \rowcolor{gray!15}\multicolumn{2}{c}{\textbf{Operator}} \\
        \midrule
        $\cdot, \times, \odot$ & Dot product, cross product, and elementwise product. \\
        $\square \| \square, \left[\square, \square, \dots, \square \right]$ & Concatenation. \\
        $\bigoplus$ & General aggregation function. \\
        $\lVert \square \rVert_2, \lVert \square \rVert_1 $  & L2 and L1 norms. L2 norm also denoted as $\lVert \square \rVert$ for brevity.  \\
        $\mathcal{S}^{(r)}_i(\square)$ & Scalarization. $\mathcal{S}^{(r)}_i(\square) = \text{flatten}(\square \cdot \mathcal{F}^{(r)}_i)$. \\

        \midrule
        \rowcolor{gray!15}\multicolumn{2}{c}{\textbf{Neural Network}} \\
        \midrule
        $d$ & Feature dimension. \\
        $\text{BN}, \text{LN}$ & Batch normalization and layer normalization. \\
        $\mathbf{W}, \mathbf{w}$ & Learnable weight matrix and learnable weight vector. \\
        $\phi, \psi, \varphi$ & General multi-layer perceptron. \\
        $\sigma$ & Activation function \\

        \bottomrule
    \end{tabular}
\end{table}
\newpage
\section{Featurization Details}\label{sec:featurization-details}
\begin{table}[htbp]
    \centering
    \caption{\textbf{Summary of protein featurization features by rank.} $[\cdot]$ means optional.}
    \label{tab:featurization-summary}
    \begin{tabular}{@{} c c l c @{}}
        \toprule
        \textbf{Rank} & \textbf{Type} & \textbf{Feature} & \textbf{Dim.} \\
        \midrule
        \multirow{7}{*}{0}
        & \multirow{5}{*}{Scalar}
        & [AA one-hot] & 23 \\
        & & 3Di one-hot & 21 \\
        & & Positional encoding & 16 \\
        & & Virtual torsion angles ($\alpha$, $\kappa$) & 4 \\
        & & Dihedral angles ($\phi$, $\psi$, $\omega$) & 6 \\
        \cmidrule(lr){2-4}
        & \multirow{2}{*}{Vector}
        & Orientation vectors & 2 \\
        & & Tetrahedral geometry & 1 \\

        \midrule

        \multirow{3}{*}{1}
        & \multirow{2}{*}{Scalar} & C$_\alpha$ distance & 1 \\
        &  & C$_\alpha$ distance positional embedding & 16 \\
        \cmidrule(lr){2-4}
        & Vector & C$_\alpha$ displacement & 1 \\

        \midrule

        \multirow{11}{*}{2}
        & \multirow{7}{*}{Scalar}
        & SSE type one-hot & 3 \\
        & & SSE size & 1 \\
        & & start/end residue positional encoding & 20 \\
        & & Consecutive SSE angles & 4 \\
        & & SSE plane torsional angle & 2 \\
        & & SSE eigenvalues & 3 \\
        & & Shape descriptors derived from eigenvalues & 5 \\
        \cmidrule(lr){2-4}
        & \multirow{4}{*}{Vector}
        & Displacement vectors between SSE start/mid/end/COM & 6 \\
        & & SSE eigenvectors & 3 \\
        & & Consecutive SSE COM displacements & 2 \\
        & & SSE (start, COM, end) $\to$ protein COM displacements & 3 \\

        \midrule

        \multirow{12}{*}{3}
        & \multirow{9}{*}{Scalar}
        & Protein size & 1 \\
        & & [AA frequency] & 23 \\
        & & SSE frequency & 3 \\
        & & SSE size mean & 3 \\
        & & SSE size std. dev. & 3 \\
        & & Protein eigenvalues & 3 \\
        & & Shape descriptors derived from eigenvalues & 5 \\
        & & Radius of gyration & 1 \\
        & & Contact density \& order & 2 \\
        \cmidrule(lr){2-4}
        & \multirow{3}{*}{Vector}
        & Global eigenvectors & 3 \\
        & & Farthest node displacements & 10 \\
        & & Nearest node displacements & 10 \\

        \bottomrule
    \end{tabular}
\end{table}

\section{GVP-TNN and ETNN Implementation Details}

\subsection{GVP-TNN}\label{subsec:gvp-tnn}
GVP-TNN is a topological enhancement of GVP-GNN~\parencite{jingLearningProteinStructure2020} that incorporates SSE-to-SSE and SSE-to-node message passing while keeping the original GVP modules unchanged. The model enables communication between different topological ranks through specialized message passing operations while maintaining E(3)-equivariance, as shown in the following equations:
\begin{align}
    \mathbf{u}^{(2)}_{i}  &= \sum_{j \in \mathcal{N}_{i}^{2 \sim 1}, (a, b) \in \mathcal{E}_{ij}^{(2)}}  \psi^{(2)} \bigl(\mathbf{h}^{(2)}_i, \mathbf{h}^{(2)}_j, \mathbf{h}^{(1)}_{ab} \bigr),\\
    \mathbf{u}^{(0)}_{i}  &= \sum_{j \in \mathcal{A}^{0 \sim 1}} \psi^{(0)}  \bigl(\mathbf{h}^{(0)}_i, \mathbf{h}^{(0)}_j, \mathbf{h}^{(1)}_{ij} \bigr), \\
    {\mathbf{h}^{(0)}_{i}}^{\prime} &= \phi^{(0)} \bigl(\mathbf{h}^{(0)}_{i}, \mathbf{u}^{(0)}_{i}, \varphi_{\text{down}}(\mathbf{u}^{(2)}_{i}) \bigr), \\
    {\mathbf{h}^{(2)}_{i}}^{\prime} &= \phi^{(2)} \bigl(\mathbf{h}^{(2)}_{i}, \sigma(\varphi_{\text{gate}}(\mathbf{u}^{(0)}_{i})) \odot \varphi_{\text{up}}( \varphi_{\text{down}}(\mathbf{u}^{(2)}_{i}) ) \bigr),
\end{align}
where $\mathcal{N}_{i}^{2 \sim 1}$ is defined as the set of 2-cells that can be connected to the 2-cell $i$ by an edge, $\mathcal{A}^{0 \sim 1}$ is the rank 0 adjacency via rank 1, $\mathcal{E}_{ij}^{(2)}$ is all the edges that can connect the 2-cell pair $(i,j)$, $\mathbf{u}^{(2)}_{i}$ and $\mathbf{u}^{(0)}_{i}$ represent the aggregated messages for rank-2 (SSEs) and rank-0 (residues) cells, respectively. The functions $\psi^{(2)}$ and $\psi^{(0)}$ are message functions that are built with GVP convolution layers, while $\phi^{(0)}$ and $\phi^{(2)}$ are update functions that are built with feedforward GVP layers. The $\varphi_{\text{down}}$, $\varphi_{\text{up}}$, and $\varphi_{\text{gate}}$ functions compress SSE information passed to residues and regulate the information flows to the next layer.

Rather than scalarizing vector features like TCPNet does, GVP-TNN only uses normalization operations when merging vector features to scalar features and processes edge messages for SSEs and residues in separate channels. This model achieves strong computational efficiency---training can be completed on a consumer GPU (NVIDIA RTX 4070 8GB) within eight hours for fold classification tasks. This balance of performance and efficiency makes GVP-TNN an ideal baseline for exploring hyperparameters that can later inform the more complex GTNN architecture.

\subsection{ETNN}\label{subsec:etnn}
ETNN~\parencite{battiloroEquivariantTopologicalNeural2024}, originally developed for small molecule representation learning, is another GTNN implementation that we adapted for protein structure analysis. Our ETNN implementation extends the basic architecture of EGNN~\parencite{satorrasEquivariantGraphNeural2021} by incorporating adjacency matrices at rank 0 via rank 2 connections in addition to the adjacency matrices at rank 0 via rank 1 connections. Unlike the original ETNN framework, which enables message passing between arbitrary ranks, this is a simplified version that focuses on demonstrating the impact of incorporating topological features while maintaining a straightforward pairwise message passing structure.

The key equations governing ETNN's message passing are:
\begin{align}
    {\mathbf{h}_i^{(0)}}^{\prime}
    &= \phi^{(0)}\Bigl(
    \mathbf{h}_i^{(0)},\,
    \sum_{r \in \{1,2\}}
    \sum_{j \in \mathcal{A}_i^{0 \sim r} }
    \phi^{(r)} \bigl(\mathbf{h}_i^{(0)}, \mathbf{h}_j^{(0)}, \|\mathbf{x}_{ij}^{(0)}\|\bigr)
    \Bigr),\\
    {\mathbf{x}_i^{(0)}}^{\prime}
    &= \mathbf{x}_i^{(0)}
    + \sum_{r \in \{1,2\}}
    \sum_{j \in \mathcal{A}_{i}^{0 \sim r} }
    \mathbf{x}_{ij}^{(0)}
    \odot \psi^{(r)} \bigl(\mathbf{h}_i^{(0)}, \mathbf{h}_j^{(0)}, \|\mathbf{x}_{ij}^{(0)}\|\bigr), \\
    \mathbf{x}_{ij}^{(0)} &= \mathbf{x}_{j}^{(0)} - \mathbf{x}_{i}^{(0)},
\end{align}
where $\mathbf{h}_i^{(0)}$ represents the scalar node features, $\mathcal{A}_{i}^{0 \sim r}$ is the adjacency matrix at rank $0$ via rank $r$, $\mathbf{x}_i^{(0)}$ represents the node coordinates, and $\phi^{(0)}$, $\phi^{(r)}$, and $\psi^{(r)}$ are simple MLPs.

\section{Alternative Frame Construction Methods}\label{sec:alternative-frames}

At node and SSE level, there are alternative ways of constructing localized frames for achieving SE(3)-equivariance beyond our edge-centric approach. Besides reducing computational cost at the expense of expressiveness, such frames can be used as vector features rather than for scalarization, where edge-centric frames remain optimal.

\textbf{Node-Level Frames:} An alternative way provided by LEFTNet~\parencite{duNewPerspectiveBuilding2023} constructs node-level frames by using the displacement between a node and its neighborhood center of mass (COM):
\begin{align}
\bar{\mathbf{x}}_i &= \frac{1}{|\mathcal{N}^{0 \sim 1}_i|}\sum_{j \in \mathcal{N}^{0 \sim 1}_i} \mathbf{x}_j \\
\mathcal{F}^{(0)}_i &= \left(
\frac{\bar{\mathbf{x}}_i - \mathbf{x}_i}{\|\bar{\mathbf{x}}_i - \mathbf{x}_i\|},
\frac{\bar{\mathbf{x}}_i \times \mathbf{x}_i}{\|\bar{\mathbf{x}}_i \times \mathbf{x}_i\|},
\frac{\bar{\mathbf{x}}_i - \mathbf{x}_i}{\|\bar{\mathbf{x}}_i - \mathbf{x}_i\|} \times \frac{\bar{\mathbf{x}}_i \times \mathbf{x}_i}{\|\bar{\mathbf{x}}_i \times \mathbf{x}_i\|}
\right)
\end{align}
where $\bar{\mathbf{x}}_i$ represents the center of mass of the neighbors of the node $i$ and $\mathcal{N}^{0 \sim 1}$ denotes a neighborhood function at rank $0$ via rank $1$.

\textbf{SSE-Level Frames:} There are two ways to construct SSE-level frames. The first method uses SSE center of mass and protein center of mass, inspired by the node-level frame used in LEFTNet, but our method uses an anchor vector (the displacement from the protein COM to the farthest residue) to avoid degeneracy:
\begin{align}
\bar{\mathbf{x}}_i &= \frac{1}{|\mathcal{B}^{2 \rightarrow 0}_i|} \sum_{j \in \mathcal{B}^{2 \rightarrow 0}_i} \mathbf{x}_j \\
\dot{\mathbf{x}} &= \mathbf{x}_{\text{argmax}_{k \in S}(\|\mathbf{x}_k\|)} \\
\mathcal{F}^{(2)}_{i} &= \left(
\frac{-\bar{\mathbf{x}}_i}{\|\bar{\mathbf{x}}_i\|},
\frac{\dot{\mathbf{x}} \times \bar{\mathbf{x}}_i}{\|\dot{\mathbf{x}} \times \bar{\mathbf{x}}_i\|},
\frac{-\bar{\mathbf{x}}_i}{\|\bar{\mathbf{x}}_i\|} \times \frac{\dot{\mathbf{x}} \times \bar{\mathbf{x}}_i}{\|\dot{\mathbf{x}} \times \bar{\mathbf{x}}_i\|}
\right),
\end{align}
where $\bar{\mathbf{x}}_i$ is the center of mass of the SSE $i$. The second method uses PCA as employed in protein-level frames, providing an alternative geometric basis for SSE orientation.

\section{Extended Experimental Details}
\label{app:details}
We use ProteinWorkshop v0.2.5~\parencite{jamasbEvaluatingRepresentationLearning2024} as our benchmarking framework, with refinements specific to our research needs. We adapt the EGNN~\parencite{satorrasEquivariantGraphNeural2021}, GCPNet~\parencite{moreheadGeometrycompletePerceptronNetworks2024}, and GVP-GNN~\parencite{jingLearningProteinStructure2020} implementation from this framework. Our featurisation scheme is inspired by~\textcite{jamasbEvaluatingRepresentationLearning2024, tanProteinRepresentationLearning2024}, including backbone-level structure features and optional amino acid sequence features depending on task needs.

For model architecture, we consistently use six layers with SiLU activations across all models, following existing works~\textcite{jamasbEvaluatingRepresentationLearning2024, tanProteinRepresentationLearning2024}. As shown in Table~\ref{tab:feature_support}, we employ 128 dimensions for scalar representations at all ranks, except for edges which use 32 dimensions to reduce computational complexity. For vector representations at each rank, we use $1/8$ of their corresponding scalar representation dimension. These architectural choices balance model capacity with computational efficiency.

For all tasks, we employ a three-layer MLP decoder with 512 embedding dimensions and ReLU activations between the layers.
Additionally, we implement an auxiliary node-level 3Di sequence denoising task following~\textcite{tanProteinRepresentationLearning2024} for all the graph-level prediction tasks to encourage effective residue-level structural representation learning even when training on graph-level objectives.
In this auxiliary task, we randomly permute half of the residues' 3Di types while masking the other half as unknown. Our preliminary results show that this auxiliary task consistently improves all model's performance across our evaluation tasks.
A two-layer MLP decoder with 128 embedding dimensions and ReLU activations is used to reconstruct the original 3Di types for all residues.

\begin{table}[ht]
    \centering
    \caption{\textbf{Model architectures and feature support across different ranks.} Checkmarks indicate support for scalar and vector features at each rank (Node, Edge, SSE, Protein), with embedding dimensions shown in the bottom row.}
    \label{tab:feature_support}
    \resizebox{0.75\textwidth}{!}{%
        \begin{tabular}{lcccccccc}
            \toprule
            & \multicolumn{4}{c}{Scalar}
            & \multicolumn{4}{c}{Vector} \\
            \cmidrule(lr){2-5} \cmidrule(lr){6-9}
            Model
            & Node & Edge & SSE & Protein
            & Node & Edge & SSE & Protein \\
            \midrule
            TCPNet-Pr
            & \checkmark & \checkmark & \checkmark & \checkmark
            & \checkmark & \checkmark & \checkmark & \checkmark \\
            TCPNet
            & \checkmark & \checkmark & \checkmark &
            & \checkmark & \checkmark & \checkmark & \\
            GVP-TNN
            & \checkmark & & \checkmark &
            & \checkmark & & \checkmark & \\
            ETNN
            & \checkmark & & \checkmark &
            & & & & \\
            GCPNet
            & \checkmark & \checkmark & &
            & \checkmark & \checkmark & & \\
            GVP-GNN
            & \checkmark & \checkmark & &
            & \checkmark & \checkmark & & \\
            EGNN
            & \checkmark & & &
            & & & & \\
            \midrule
            Emb. Dim.
            & 128 & 32 & 128 & 128
            & 16 & 4 & 16 & 16 \\
            \bottomrule
        \end{tabular}%
    }
\end{table}

Training was performed on a single NVIDIA A100 80GB GPU with a batch size of 32 and the Adam optimizer (no weight decay) with an initial learning rate of 0.001.
The learning rate was dynamically adjusted using a ReduceOnPlateau scheduler, which decreased the rate by a factor of 0.6 when the validation metric showed no improvement for 5 consecutive epochs.
Models were trained until convergence, with training stopped if any of the following criteria were met: (1) reaching 150 epochs, (2) exceeding task-specific time limits (inverse folding: 4 hours, fold classification: 4.5 hours, gene ontology: 10 hours, antibody developability: 2 hours), or (3) triggering early stopping after 10 epochs without validation improvement.
These time limits were carefully chosen through preliminary experiments to allow all models to reach convergence or at least near convergence while staying within our computational budget.
Due to computational constraints, we were unable to perform extensive hyperparameter tuning, try various feature combinations, or run multiple trials for each experiment.
However, we tried to keep our settings close to existing benchmark experiments~\parencite{jamasbEvaluatingRepresentationLearning2024} and ensured consistent training conditions across all model variants to enable fair comparisons.

\section{Dataset Details and Task-Specific Implementation}\label{sec:dataset-details-and-task-specific-implementation}

\textbf{CATH 4.4 (Inverse Folding):} We use the CATH 4.4 dataset~\parencite{knudsenTheCATHDatabase2010, wamanCATHV44Major2025} containing 34,652 protein structures (32,652/1,000/1,000 train/validation/test split) filtered at 40\% sequence similarity. This node-level multi-class classification task predicts amino acid sequences that would fold into given 3D protein structures. Each backbone position must be classified into one of 23 amino acid classes. The task is evaluated using perplexity (measuring model uncertainty) and accuracy (percentage of correctly predicted amino acids). The 40\% sequence similarity filtering prevents information leakage between splits while maintaining structural diversity.

\textbf{SCOP 1.75 (Fold Classification):} We employ the SCOP 1.75 dataset~\parencite{murzinSCOPStructuralClassification1995} for graph-level classification of proteins into 1,195 distinct fold classes based on 3D structural arrangements. The dataset contains 12,312 training and 736 validation structures, with three difficulty levels for testing: Family (1,272 proteins, no constraints), Superfamily (1,254 proteins from different families), and Fold (718 proteins from different superfamilies). The Fold split represents the greatest challenge due to structural similarity constraints, requiring models to generalize learned fold patterns to unseen superfamilies. Performance is measured by classification accuracy across all three difficulty levels.

\textbf{Gene Ontology (Cellular Component Prediction):} This graph-level multi-label classification task using 16,019/1,294/1,714 training/validation/test structures with <30\% sequence similarity between training and test sets. The dataset, curated from experimental PDB structures, uses the standardized Gene Ontology framework for protein localization prediction. Due to computational constraints, we focus specifically on the cellular component ontology rather than all three GO categories (molecular function, biological process, cellular component). Performance is evaluated using maximum F1 score (F1\textsubscript{max}) at optimal classification threshold.

\textbf{SabDab (Antibody Developability):} We use the SabDab database~\parencite{dunbarSAbDabStructuralAntibody2014} containing 2,388 antibody structures (1,669/241/478 train/validation/test split) for binary classification of therapeutic viability. This task predicts whether antibodies possess favorable physicochemical properties for drug development based on developability index metrics. The structures comprise both heavy and light chains, making them substantially larger than proteins in other datasets. Due to class imbalance, performance is evaluated using Area Under Precision-Recall Curve (AUPRC). The task requires understanding antibody-specific structural features and evaluates model performance specifically on the immunoglobulin fold.

\section{Topological Deep Learning Primer}\label{sec:tdl-primer}

This section provides essential resources and foundational concepts for researchers new to Topological Deep Learning. TDL extends traditional neural networks to operate on complex, hierarchical data structures beyond simple graphs. While conventional deep learning methods process flat data representations such as images or sequences, TDL addresses multi-scale structures where relationships exist simultaneously across multiple hierarchical levels, analogous to processing linguistic information at word, sentence, paragraph, and document levels concurrently.

Topological domains represent mathematical frameworks for complex data structures that generalize standard graphs to capture higher-order relationships. These domains extend pairwise connectivity to multi-dimensional organizational structures. Combinatorial complexes provide flexible topological representations that unify hierarchical organization with set-type relationships while removing the strict boundary constraints of simplicial complexes. Topological neural networks enable message passing across arbitrary ranks within topological structures, facilitating information flow both within and between hierarchical levels.

For comprehensive understanding, readers should begin with \textcite{papillonArchitecturesTopologicalDeep2023}, which provides a thorough overview of TDL domains and neural network architectures. The mathematical framework for combinatorial complexes is established in \textcite{hajijTopologicalDeepLearning2022}. For geometric applications relevant to molecular systems, ETNN by \textcite{battiloroEquivariantTopologicalNeural2024} demonstrates equivariant topological neural networks operating on combinatorial complexes.

Practical implementation resources include the TopoX library~\parencite{hajijTopoXSuitePython2024}, a Python framework for experimenting with topological domains and neural networks, and TopoBench~\parencite{telyatnikovTopoBenchFrameworkBenchmarking2025}, a standardized benchmarking suite for evaluating topological neural network performance.

\end{document}